\documentclass{article}

\usepackage{graphicx}
\usepackage[utf8]{inputenc} % allow utf-8 input
\usepackage[T1]{fontenc}    % use 8-bit T1 fonts
\usepackage[hidelinks]{hyperref}       % hyperlinks
\usepackage{url}            % simple URL typesetting
\usepackage{booktabs}       % professional-quality tables
\usepackage{amsfonts}       % blackboard math symbols
\usepackage{nicefrac}       % compact symbols for 1/2, etc.
\usepackage{microtype}      % microtypography
\usepackage{lipsum}
\usepackage{amsmath}
\usepackage{setspace}

\usepackage[accepted]{ecopadl2020}

\icmltitlerunning{Really Useful Synthetic Data}
\usepackage{caption}
\usepackage{subcaption}

\begin{document}

\twocolumn[
\icmltitle{Really Useful Synthetic Data \\ A Framework to Evaluate the Quality of Differentially Private Synthetic Data}



\icmlsetsymbol{equal}{*}

\begin{icmlauthorlist}
\icmlauthor{Christian Arnold}{equal,af1}
\icmlauthor{Marcel Neunhoeffer}{equal,af2}

\end{icmlauthorlist}

\icmlaffiliation{af1}{Cardiff University}
\icmlaffiliation{af2}{University of Mannheim}

\icmlcorrespondingauthor{Christian Arnold}{arnoldc6@cardiff.ac.uk}

\icmlkeywords{Machine Learning, ICML, Differential Privacy}

\vskip 0.3in
]



\printAffiliationsAndNotice{\icmlEqualContribution} % otherwise use the standard text.

%\title{Really Useful Synthetic Data -- Promises and Challenges of Releasing Sensitive Information With Differentially Private Data Synthesizers}
%\title{Really Useful Synthetic Data \\ \large{A Framework to Evaluate the Quality of Differentially Private Synthetic Data}}

\begin{abstract}

Recent advances in generating synthetic data that allow to add principled ways of protecting privacy -- such as Differential Privacy -- are a crucial step in sharing statistical information in a privacy preserving way. But while the focus has been on privacy guarantees, the resulting private synthetic data is only useful if it still carries statistical information from the original data. To further optimise the inherent trade-off between data privacy and data quality, it is necessary to think closely about the latter. What is it that data analysts want? Acknowledging that data quality is a subjective concept, we develop a framework to evaluate the quality of differentially private synthetic data from an applied researcher's perspective. Data quality can be measured along two dimensions. First, quality of synthetic data can be evaluated against training data or against an underlying population. Second, the quality of synthetic data depends on general similarity of distributions or on performance for specific tasks such as inference or prediction. It is clear that accommodating all goals at once is a formidable challenge. We invite the academic community to jointly advance the privacy-quality frontier.

\end{abstract}

% keywords can be removed
% \keywords{Synthetic Data \and Differential Privacy \and Data Quality}

\iffalse\textit{Previous versions have been presented at the Digital Governance Research Colloquium of the Hertie School, the Workshop ``(Differentially) Private Synthetic Data'' at Cardiff University, at the 3rd Social Science Foo Camp, Menlo Park and at the 1st OpenDP Community Meeting 2020. Comments are welcome. Many thanks already for helpful comments to Thomas Gschwend, Frauke Kreuter and Sebastian Sternberg. Chris is grateful for support through Cardiff's ``Darlithwyr Disglair Development Scheme''. Marcel benefited tremendously from the ``Data Privacy: Foundations and Applications'' program at the Simons Institute for the Theory of Computing, UC Berkeley.}\fi

% == Short Version Paper ===========================================

%\setcounter{page}{1}
\begin{Large}
\centering Summary for EcoPaDL at ICML 2020
\end{Large}

\section{Introduction}

% The problem to be solved
Humanity is collecting data at an unprecedented level -- and very often this data is sensitive. Scientific studies across disciplines often rely on private information. Government agencies hold confidential data about their citizens. And companies -- think Facebook, Google or Twitter -- are recording individuals' (online) behavior. Ideally, all this data could be shared: Scientists could draw from the same sources to jointly advance knowledge; citizens could use public sector data to hold governments accountable; and companies could externalise data services. However, important privacy concerns do not allow to freely circulate all these records.

% This is how we can solve it with existing tools
Is it really necessary to share the data itself? Most data users care about the statistical information in the data and not so much about learning sensitive traits. A possible solution for sharing such statistical information is synthetic data: an artificial copy of the original data-set that ideally shares the same statistical properties. Analysts can work with synthetic data as if it were the original data and reach similar conclusions. In addition, synthetic data introduces a first layer of privacy protection. The individual records are completely synthetic and do not exist in real life. Nevertheless, synthetic data can still reveal sensitive information. If a strong outlier -- say e.g. Bill Gates -- is in the original data, the laws of statistics dictate that the synthetic version of the data reproduce his record almost perfectly. Just like any other data, synthetic data requires privacy protection, ideally in a principled and future proof way. Differential Privacy is such a privacy standard and it has recently attracted a lot of attention in academia, business and government agencies alike \citep{abowd2019, Dwork2006}. It offers mathematically proven privacy guarantees without making assumptions about the distribution of the original data, available side knowledge or computational power. Differentially private synthetic data brings both of these concepts together and allows to share statistical information with privacy guarantees.

But synthetic data needs to be not only private, it also needs to be useful. The ultimate goal of this paper is therefore to improve the methods that can release high-quality differentially private synthetic data. Differentially private data synthesizers protect privacy by adding noise to the original statistical information. But the noise runs the risk of blurring this information. Algorithms to synthesize differentially private data have become increasingly available and they perform well in pilot studies with benign (toy-)data. Often, this work still fails to live up to the challenges data users typically face: in the wild, micro data\footnote{``Micro data are unit-level data obtained from sample surveys, censuses and administrative systems. They provide information about characteristics of individual people or entities such as households, business enterprises, facilities, farms or even geographical areas such as villages or towns. They allow in-depth understanding of socio-economic issues by studying relationships and interactions among phenomena. Micro data are thus key to designing projects and formulating policies, targeting interventions and monitoring and measuring the impact and results of projects, interventions and policies.'' \cite{Benschop2019}} can contain discrete and continous attributes; it can have missing data; certain combinations of variables might not make a lot of sense -- e.g. pregnant men; or data structures can be nested such as individuals in groups.

To advance the field, it is necessary to better understand the trade-off between data privacy and data quality. While a lot of attention rightly focused on establishing strong privacy guarantees, the next step has to consider the resulting data quality more closely \citep{abowd2019c, elliot2018, hundepool2012}. We therefore formulate a user-centric framework that measures the quality of synthetic data along two dimensions. First, while in some cases data needs to be similar to training data, in other cases the underlying 'unknown' population data is playing a more vital role. Second, quality can be expressed in general terms with regards to the overall distribution, or in specific terms to address two main interests of the applied data user -- inference and prediction.\footnote{Even though our framework is specifically designed for the purpose of private synthetic data, it generalises well to other forms of synthetic data, too.} Building on scenarios that describe applied researchers' main data related obstacles, we suggest two different 'disciplines' in which a differentially private data synthesizer should perform well: resulting synthetic data needs to be useful at different levels of privacy guarantees and also at different training set sizes. We maintain a dedicated homepage to keep track of the performance of differentially private data synthesizers\iffalse\footnote{\url{http://privatesyntheticdata.github.io}}\fi.

Our paper addresses several audiences. First, we turn to applied researchers who typically use micro data and offer an overview about state-of-the art possibilities of generating differentially private synthetic data. Second, we reach out to those with a rigorous understanding about how to develop differentially private data synthesizers. In spotlighting current limitations, we intend to point to avenues for future research. Our hope is to ultimately spark a fruitful conversation between applied researchers, Statisticians and Computer Scientists about really useful privacy preserving synthetic micro data.

\section{Core Requirements for a Useful DP Data Synthesizer}
% These are the problems that have to be taken into account
% - DP Synthetisation from a statistical point of view
% - Typical Data Structures
% - Application Scenarios

Differentially private GANs (DP-GANs) were the first class of models to synthesize differentially private data sets from different domains and for different uses. In existing studies, quality is either assessed by looking at the visual quality of generated images \citep[e.g.][]{Xie2018, FrigerioOGD19, TKP20} or the classification evaluation of a classification/prediction task \citep[e.g.][]{Beaulieu-Jones2017}. Yet, there is no clear definition of what data quality actually implies beyond these measures. In the absence of natural measures, it is more difficult to assess the overall quality of synthetic data.

The question what a useful DP data synthesizer actually is depends on the user. From the perspective of a data analyst who ultimately has to work with synthetic data, there are three core challenges. First, the synthetization process has to preserve statistical properties. Second, the data synthesizer has to be able to live up to the challenge of the realities of applied researchers' data. Ultimately, DP data synthetization needs to occur in typical scenarios of different sample sizes and different privacy requirements.

\subsection{Differentially Private Data Synthetization from a Statistical Viewpoint}
% Inference

Applied researchers may have different expectations when they analyse private synthetic data. Some are interested in rather descriptive analyses of their data. They consider their data not as a sample from what potentially could be a larger population of data sets. This is the case whenever the original data captures a unique data source. For example, studies on political elites such as candidate studies for elections comprise all available data; or data on banking transactions that help identify money laundering are typically also unique instances. Data users analyse this kind of data with a descriptive goal in mind. They do not assume a larger population behind this data about which they want to draw inferences. All their analyses run with differentially private synthetic data should ideally lead to similar results as the analyses using the original data.

In other cases, analysts may want to do more. The training data the DP synthesizer has to learn from is only a sample of many more possible data sets drawn from a larger population. The analyst might be interested in estimating the value of population parameters. In a similar vein, out of sample prediction works well if a model trained on differentially private synthetic data is capable of generalising beyond the training set itself. The model then has to be able to capture information about the process that generates data in the population in the first place.

Usually, inference means drawing a random sample from the population to learn more about a true unknown parameter $\theta$. Statistics provides us with estimators for the parameter $\hat{\theta}$ and the appropriate variance $V(\hat{\theta})$. However, in the case of differentially private synthetic data this process looks different. Synthesizing differentially private data turns inference into a multi-phased problem \citep{Xie2017}: The researcher now has to infer the true underlying population parameter $\theta$ on the basis of the synthesized and privatized data which was itself generated from a random sample from the population -- both in the estimate $\hat{\theta}$ and in the estimate of $\hat{V}(\hat{\theta})$, too. Conceptually, these sources of uncertainty are distinct.

% congeniality:
Valid statistical inference on private synthetic data is possible if the data synthetization process is unbiased and consistent. This is typically the case when the model for synthetization and the model for subsequent analyses are congenial \citep{Meng1994}. While congenial differentially private data synthesizers do exist -- and allow to exactly specify bias and efficiency of the private synthetic data \citep[e.g.][]{McClure2012} -- the reality often looks different. Congeniality can only be achieved if the data curator succeeds in anticipating and hence nesting (all) models of the analyst.

\citet{Xie2017} propose a general multi-phase inference framework in the face of uncongeniality. It is not enough if each, the data synthesizer and the analyst, use statistically valid models for themselves. Both models need to be congenial to allow for inferring the true unknown data generating process. If they are uncongenial, inference on the basis of the synthetic data may be unbiased, but it may lead to wrong conclusions about the efficiency of the estimate. Two scenarios are possible: If the synthesizer model is more saturated than the analyst's model, insights on multiple imputation \citep{Rubin1993, Xie2017} suggest that the result is confidence proper. However, in the second scenario -- if the synthesizer model is less saturated than the analyst's model -- super efficiency leads to too narrow confidence bounds. \citet{Xie2017} show how to be better save than sorry and prove upper uncertainty bounds: Doubling the variance estimate is enough to generate a conservative estimate for multi-phase inference even under uncongenial models.

\subsection{Applied Researcher's Typical Data}

% The different challenges
Applied researchers' micro data have a number of characteristics that make it challenging to generate differentially private synthetic data for them. Indeed, data can be continuous. More often, however, micro-data is discrete. Categories could be ordinal or without any particular natural order to them. Structural zeros preclude logically impossible values such as pregnant three-year-olds. Often, parts of the data are missing -– either at random or not at random. Finally, data structures can be nested. In time series data, the same individual might be observed at different points in time. In hierarchical data, individuals are part of a larger group that comes with its own measurements, such as students in a classroom.

\subsection{Useful Data for Different Sample Sizes and Privacy Levels}
Finally, a data synthesizer should be able to rise to these challenges in two different \emph{disciplines}. Differentially private data synthesizers ideally perform well at different levels of privacy protection. The literature considers values up to $\epsilon=1$ acceptable. A value of $\epsilon=5$ is already quite a considerable privacy leak, but, depending on the data, still worthwhile exploring. In addition, the algorithms need to perform at different sample sizes. Applied researchers' sample sizes range from clinical studies ($500$ individuals) to surveys with $10,000$ individuals. Data sets with more than $100,000$ individuals do exist, but are fairly rare.
% \iffalse\footnote{After all, ``there is essentially only one scientific way to evaluate and compare statistical procedures -- show how they work when applied repeatedly, in reality, via simulation or in thought experiments.\citep{Xie2017}''}\fi

\section{A General Framework to Evaluate the Quality of Private Synthetic Data}

In a perfect world, a differentially private data synthesizer should work well for any data challenges and across the different \emph{disciplines}. We now turn to what we mean by ``working well''.

Data quality is context dependent. Measures for the quality of differentially private synthetic data should reflect the needs of those who work with this kind of data. We suggest to distinguish between quality measures along two dimensions.%\iffalse The first dimension is concerned with the similarity of the differentially private synthetic data $x^{*}$.\fi
The first dimension distinguishes between the similarity of the differentially private synthetic data to the training data (i.e. the data that was used to generate the synthetic data) and the generalization properties of the differentially private synthetic data. On the one hand, data users might be particularly concerned with the similarity of differentially private synthetic data to the original training data. On the other hand, data users might care less about similarity of the data set to one particular training set. Instead, data analysts may be interested in inferring information about the underlying population -- to either identify relationships between variables or build models for out-of-sample predictions.

The second quality dimension is concerned with general data quality and specific data quality \citep{Snoke2018}\footnote{Note that \citet{Snoke2018} speak of general and specific utility.}. General data quality describes the overall distributional similarity between the real data and synthetic data sets. In turn, to reflect specific analyses data analysts expect to run on synthetic data we introduce specific data quality. This could be the similarity of marginal distributions, the performance of regression models or the predictive performance of machine learning models trained on synthetic data sets as compared to the same analysis carried out on the real data set.
% TODO add footnote to point out relative perspective of marginal similarities. Define general and specific utility better and in line with Snoke et al.

% What data utility measures to chose from
% Grammar when explaining: a) what do you want to do b) how does it work c) what is a good and a bad value
Table~\ref{tab:utility} provides an overview of data quality measures along both dimensions, resulting in overall four different quadrants. The ultimate choice of the quality measures is indeed subjective and needs to fit their purpose \citep{Shlomo2018}. We first turn to data quality of private synthetic data relative to the training data in the top left cell of Table~\ref{tab:utility}.

\begin{table*}[!htbp]
\centering
\caption{Proposed Quality Measures for Private Synthetic Data Along Two Dimensions.}
\label{tab:utility}
\begin{tabular}{@{}p{2.3cm}p{5.3cm}p{5.3cm}@{}}
                           & \textbf{General Data Quality}                  & \textbf{Specific Data Quality} \\
\midrule
\textbf{Training Data Similarity} & Similarity of DP synthetic data and sample 1-way marginals: Wasserstein Randomisation Test         & Compare $\hat\theta^{*}$ with $\hat\theta$: average percent bias (difference between coefficients calculated on synthetic data and training data)                \\
\addlinespace
                           & $pMSE$ bewteen synthetic data and training sample                 & Compare $\hat V^{*} \left(\hat\theta^{*}\right)$ with $\hat V \left(\hat\theta\right)$: average variance and covariance ratio \\
% \addlinespace
                        %   &         &  Prediction RMSE w.r.t. sample           \\
\midrule
\textbf{Generalisation Similarity}   &  Similarity of DP synthetic data and independent test sample 1-way marginals: Wasserstein Randomisation Test  & Compare $E\left(\hat\theta^{*}_{m}\right)$ with $\theta$: average percent bias (difference between true population values and coefficients on synthetic data)\\
\addlinespace
                           & $pMSE$ bewteen synthetic data and independent test samples     & Compare $E\left(\hat V^{*}_{m} \left(\hat\theta^{*}_{m}\right)\right)$ with $V\left(\theta\right)$: coverage, average width \\
\addlinespace
                           &  & Prediction RMSE w.r.t. further samples from population             \\

\end{tabular}
\end{table*}

To compare the overall similarity of the joint distribution and to calculate the general quality of the synthetic copies we rely on the propensity score mean squared error (pMSE) ratio score for synthetic data as proposed by \citet{Snoke2018}. The pMSE requires training a discriminator (e.g. a CART model) that is capable to distinguish between real and synthetic examples. A synthetic data set has high general data quality, if the model cannot distinguish between real and fake examples. It is possible to approximate the pMSE by resampling the null distribution of the pMSE. The pMSE ratio score divides the pMSE with the expectation of the null distribution. A pMSE ratio score of 0 would indicate that the synthetic and real data are identical.
% \iffalse Next, to compare the overall similarity of the joint distribution, we rely on the propensity-score mean-squared error ($pMSE$) \citet{Snoke2018}. Their procedure first match data from the original and the training set and then empirically investigate to which extent powerful prediction models (e.g. CART) are capable of distinguishing whether an observation originates from the training data set or the synthetic data set. A ratio $pMSE = 1$ indicates correct synthesis of the data. \fi
% Eventually: $pMSE$ longer. Or Wasserstein Distance
One of the most relevant concerns when releasing differentially private synthetic data remains the similarity of $1-way$ marginals. The Wasserstein distances measures the difference between two univariate distributions with unknown statistical properties \citep{Robinson2017}. To compare this data specific metric across different attributes, it has to be put into relative perspective: We propose a metric based on the idea of randomization inference \citep{Basu1980, Fisher1935}. Each data point is randomly assigned to one of two data sets and the similarity of the resulting two distributions is measured with the Wasserstein distance. Repeating this random assignment a great number of times (e.g. $10'000$ times) provides an empirical approximation of the distances' null distribution. Similar to the pMSE ratio score we then calculate the ratio of the measured Wasserstein distance and the median of the null distribution to get a Wasserstein distance ratio score that is comparable across different attributes. Again a Wasserstein distance ratio score of 0 would indicate that two marginal distributions are identical. Larger scores indicate greater differences between distributions. %\iffalse\footnote{Please note the similarity to randomization inference \citep{Basu1980, Fisher1935}.}  How often is the Wasserstein distance between the original data and the synthetic data larger than between two randomly assigned data sets? The resulting proportion can be interpreted as a $p$-value, and we suggest to follow the convention. If $p > 0.05$, then the two marginal distributions are very close and the data quality is high. Instead, if $p < 0.05$, both marginals are fairly distinct, and hence data quality is low. To measure the overall similarity between original data and synthetic data in its margins, we calculate the proportion of Wasserstein distances larger than the p-value $pWD_{1d}$ for all $1-way$ marginals, and $pWD_{2d}$ for all $2-way$ marginals.\fi

The similarity between the private synthetic data and the training data addresses data analysts' specific requirements for data quality.\footnote{For similar measures used in multiple imputation, see \citet{vanBuuren2018}.} They might first be interested in understanding correlations -- and multivariate statistical models quantify these correlations while controlling for the influence of other variables. It is straightforward to calculate bias of parameters $\hat \theta^{*}$ estimated on synthetic data $x^{*}$ in comparison to $\hat \theta$ estimated on $x$ as percent bias with their normalised difference. We then aggregate all biases to one score by taking their average. To compare the variances, we follow \citet{Drechsler2018}, and measure the ratio between the variances of the parameters $\hat V^{*} \left(\hat\theta^{*}\right)$ in the private synthetic data and divide it with the variances $\hat V \left(\hat\theta\right)$ of the original training data. The resulting variance ratio equals 1 if they are both the same. Averaging across all ratios expresses their general difference. % Finally, the RMSE allows to compare the predictive performance of the models $m^{*}$ learned on the data $x^{*}$. Simply use the model $m^{*}$ to predict outcomes using data from the unseen original data $x$ -- and tell the difference using the RMSE.

In other cases, analysts are more interested in using samples to generalise beyond the observations in the training set and make inferences about a larger population. When using private synthetic data in this setting, the data analyst needs to account for three layers of random errors in this data, first sampling error when drawing the training data from the underlying population, second model uncertainty when training the data synthesizer (even without Differential Privacy) and third the privacy noise that is introduced to the training of the data synthesizer to release differentially private synthetic data. We suggest to empirically validate data quality in a Monte Carlo framework and average results across $1'000$ differentially private synthetic data sets.\footnote{We chose the sample size for pragmatic reasons. Some of the differentially private data synthesizers may take non-negligible time for training. In the light of today's available computational power we acknowledge that $1'000$ data sets may be not enough to complete integrate over all uncertainty from the three respective steps. We chose $10^3$ data sets in line with what \citet{Rubin1996} considered sufficient for imputation.} They are the result of $l = 10$ different training data sets -– with another $10$ independent data sets held out as test sets. Then, train $m = 10$ independent data synthesizers on each of the training sets and finally release $n = 10$ data sets from each of these synthesizers.

The first relevant measure for general data quality is the similarity of marginals, again calculated with the Wasserstein distance ratio score. This time, however, we pair each private synthetic data set with the respective held out data sets to understand the capacity to generalize. %\iffalse We calculate the overall similarity of the synthetic data with the held out data and count the overall percentage of data sets that fall below a similarity threshold of $p < 0.05$. As before, there is one measure for all 1-way marginals and another overall measure for $2-way$ marginals.\fi
In a similar vein, the $pMSE$ ratio score again summarizes the overall similarity between joint distribution. We calculate it against the hold out test sets and summarize by averaging across all $pMSE$ ratios.

For the specific data quality measures, the analyst might be concerned with bias of the parameters $\hat \theta^{*}_{i}$ of a multivariate regression model. We calculate bias for each of the synthetic data sets, comparing it to the values that define the true underlying data generating process. Standardising and then averaging across all parameters gives an overall percentage value that summarises bias in general. The coverage rate of the variance captures efficiency \citep{Drechsler2018}. How often does the 90\% confidence interval include the true value from the population? Averaging across all samples and cases yields one single measure. Finally, an analyst might be interested in generating a model that is capable of predicting values for out-of-sample prediction. Here, we estimate a model on each of the synthetic datasets. We then predict values for unseen data in the held out validation sets and summarise all results with an average RMSE.

% \iffalse
% Figure \ref{fig:analysisplan1} provides two charts that summarise the procedure for training data similarity (Figure~\ref{fig:training_data_similarity}) and generalisation similarity (Figure~\ref{fig:generalisation_similarity}).

% \fi

% \begin{figure}
%     \centering
%     \includegraphics[width = .35\textwidth]{figures/data_challenge.png}
%     \caption{}
%     \label{fig:analysisplan1}
% \end{figure}

\subsection{The Data Challenge}
\begin{figure}
     \centering

         \includegraphics[width=0.8\columnwidth]{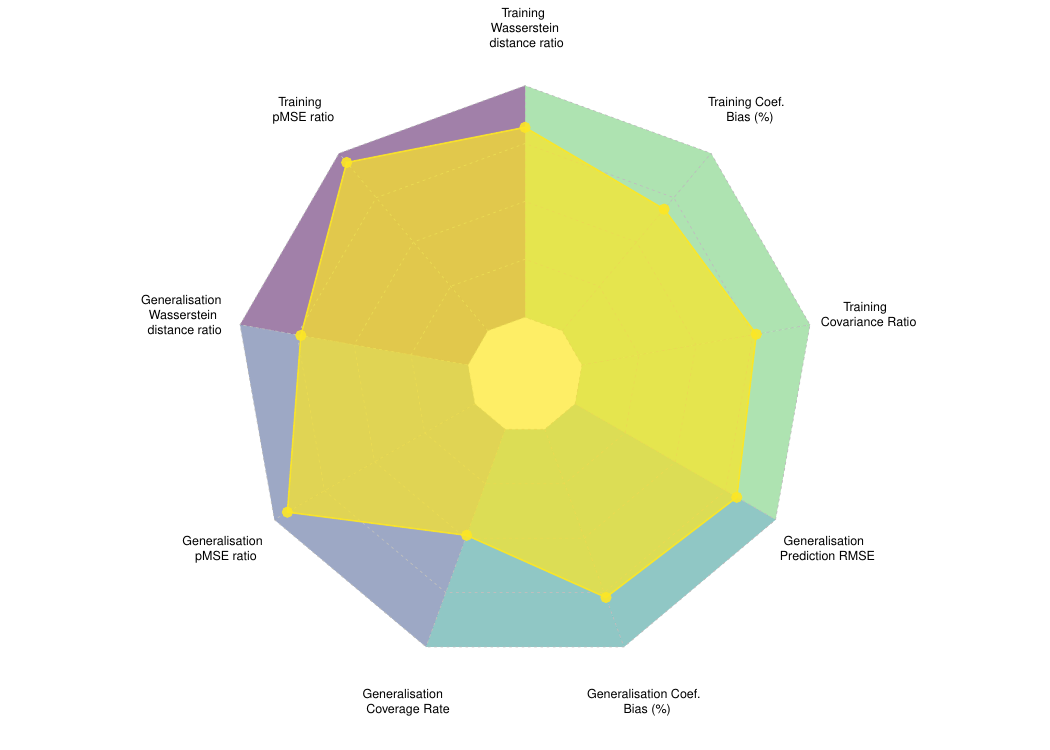}
         \caption{Summary Chart for Data Quality.}
         \label{fig:summary}

\end{figure}

% Doing it
We envision a data challenge that tests differentially private data synthesizers for potentially different data scenarios (e.g. the baseline scenario, missing data, nested data) in the two related disciplines (training data size and privacy budget). For the challenge, we measure data quality according to Table \ref{tab:utility}. For a quick comparison of different differentially private data synthesizers we summarize the measures in an easy to read chart (see Figure \ref{fig:summary}). In this hypothetical example, the yellow polygon displays the performance of a differentially private data synthesizer for the nine measures from above. The further out (the bigger) the edges of the polygon, the better the performance of a synthesizer in that domain. With this chart it is also easy to see whether a particular data synthesizer does well for some measures of data quality, while it fails for others. The ultimate goal should be to balance the performance for all measures.

To document the progress of differentially private data synthesizers, we plan to maintain an internet repository at \url{privatesyntheticdata.github.io}. We will also use the homepage to describe the data challenge more fully. You can learn more about the details for replication, i.e. a summary of the exact specifications for each scenario and each of the three ``disciplines''. We understand that researchers who submit a new algorithm do not undermine the challenge, e.g. by tailor suiting the algorithm to the particular \emph{ex-ante} known data generating process. We also trust that any hyperparameter optimization occurs without explicitly overfitting in a non-private way.

% \section{Application?}
\section{Conclusion}

Building on scenarios that describe applied researchers' main data related obstacles we seek to contribute to the development of differentially private data synthesizers and propose a joint and unified benchmark. It is our hope that by formulating a problem that needs to be solved, we offer a focus point for further collaboration across disciplines. Our suite of well defined data generating processes illustrates typical challenges of applied researchers' data, i.e. discrete data, structural zeros, missing data and nested data. Data generating algorithms should be able to account for these challenges in two different settings: at different levels of privacy and at different amounts of available training data. Finally, to assess the quality of the resulting synthetic data from the vantage point of applied researchers we introduce a set of appropriate data quality measurements along two dimensions: On the one hand similarity towards the training data or the underlying population, on the other hand general and specific data quality. All differentially private data synthesizers should ideally be capable of rising to these challenges under different settings: at different levels of privacy and with different training data sizes. They ideally also synthesize data that is similar to the training sample and generalizes well to the underlying population.
Differentially private synthetic data contributes to solving important current challenges. It offers principled privacy guarantees, opening the door to sharing data freely. Access to the statistical information in all the data currently collected promises to open the door wide to a great range of positive externalities -- be they in science, government or business.

% == Complete Paper ===========================================
\clearpage

\setcounter{section}{0}
\setcounter{page}{1}

\twocolumn[
\icmltitle{Really Useful Synthetic Data \\ A Framework to Evaluate the Quality of Differentially Private Synthetic Data}




\vskip 0.3in
]




%\title{Really Useful Synthetic Data -- Promises and Challenges of Releasing Sensitive Information With Differentially Private Data Synthesizers}
%\title{Really Useful Synthetic Data \\ \large{A Framework to Evaluate the Quality of Differentially Private Synthetic Data}}

\begin{abstract}

Recent advances in generating synthetic data that allow to add principled ways of protecting privacy -- such as Differential Privacy -- are a crucial step in sharing statistical information in a privacy preserving way. But while the focus has been on privacy guarantees, the resulting private synthetic data is only useful if it still carries statistical information from the original data. To further optimise the inherent trade-off between data privacy and data quality, it is necessary to think closely about the latter. What is it that data analysts want? Acknowledging that data quality is a subjective concept, we develop a framework to evaluate the quality of differentially private synthetic data from an applied researcher's perspective. Data quality can be measured along two dimensions. First, quality of synthetic data can be evaluated against training data or against an underlying population. Second, the quality of synthetic data depends on general similarity of distributions or on performance for specific tasks such as inference or prediction. It is clear that accommodating all goals at once is a formidable challenge. We invite the academic community to jointly advance the privacy-quality frontier.

\end{abstract}

\begin{Large}
\centering Full Paper
\end{Large}

%\onehalfspacing
%==============================================================================================================
\section{Introduction}

% The problem to be solved
Humanity is collecting data at an unprecedented level -- and very often this data is sensitive. Scientific studies often rely on private information. Government agencies hold confidential data about their citizens. And companies -- think Facebook, Google or Twitter -- are recording individuals' (online) behavior. Ideally, all this data could be shared: Scientist could draw from the same sources to jointly advance knowledge; citizens could use public sector data to hold governments accountable; and companies could externalise data services. However, important privacy concerns do not allow to freely circulate all these records.

% This is how we can solve it with existing tools
Is it really necessary to share the data itself? At the end of the day, most data users care about the statistical information in the data -- and not so much about learning sensitive traits. A possible solution is synthetic data: an artificial copy of the original data-set that ideally shares the same statistical properties. Analysts can work with synthetic data as if it were the original data and reach similar conclusions. In addition, synthetic data introduces a first layer of privacy protection. The individual records are completely synthetic and do not exist in real life. Nevertheless, synthetic data can still reveal sensitive information. If a strong outlier -- say e.g. Bill Gates -- is in the original data, the laws of statistics dictate that the synthetic version of the data reproduce his record almost perfectly. Just like any other data, synthetic data requires privacy protection, ideally in a principled and future proof way. Differential Privacy is such a privacy standard and it has recently attracted a lot of attention in academia, business and government agencies alike \citep{abowd2019, Dwork2006}. It offers mathematically proven privacy guarantees without making assumptions about the distribution of the original data, available side knowledge or computational power. Differentially private synthetic data brings both of these concepts together and allows to share statistical information with privacy guarantees.

But synthetic data needs to be not only private, it also has to be useful. The ultimate goal of this paper is therefore to improve the methods that can release high-quality differentially private synthetic data. Differentially private data synthesizers protect privacy by adding noise to the original statistical information. But the noise runs the risk of blurring this information. Algorithms to synthesize differentially private data have become increasingly available and they perform well in pilot studies with benign (toy-)data. Often, this work still fails to live up to the challenges data users typically face: in the wild, micro data\footnote{``Micro data are unit-level data obtained from sample surveys, censuses and administrative systems. They provide information about characteristics of individual people or entities such as households, business enterprises, facilities, farms or even geographical areas such as villages or towns. They allow in-depth understanding of socio-economic issues by studying relationships and interactions among phenomena. Micro data are thus key to designing projects and formulating policies, targeting interventions and monitoring and measuring the impact and results of projects, interventions and policies.'' \cite{Benschop2019}} is mostly discrete; it can have missing data; certain combinations of variables might not make a lot of sense -- e.g. pregnant men; or data structures can be nested such as individuals in groups.

To advance the field, it is necessary to better understand the trade-off between data privacy and data quality. While a lot of attention rightly focused on establishing strong privacy guarantees, the next step has to consider the resulting data quality more closely \citep{abowd2019c, elliot2018, hundepool2012}. We therefore formulate a user-centric framework that measures the quality of synthetic data along two dimensions. First, while in some cases data needs to be similar to training data, in other cases the underlying 'unknown' population data is playing a more vital role. Second, quality can be expressed in general terms with regards to the overall distribution, or in specific terms to address two main interests of the applied data user -- inference and prediction.\footnote{Even though our framework is specifically designed for the purpose of private synthetic data, it generalises well to other forms of synthetic data, too.} Building on scenarios that describe applied researchers' main data related obstacles, we suggest two different 'disciplines' in which a differentially private data synthesizer should perform well: resulting synthetic data needs to be useful at different levels of privacy guarantees and also at different training set sizes. We maintain a dedicated homepage to keep track of the performance of differentially private data synthesizers\footnote{\url{http://privatesyntheticdata.github.io}}.

Our paper addresses several audiences. First, we turn to applied researchers who typically use micro data and offer an overview about state-of-the art possibilities of generating differentially private synthetic data. Second, we reach out to those with a rigorous understanding about how to develop differentially private data synthesizers. In spotlighting current limitations, we intend to point to avenues for future research. Our hope is to ultimately spark a fruitful conversation between applied researchers, Statisticians and Computer Scientists about really useful privacy preserving synthetic micro data.

% TODO quality and utility

\section{Sharing Statistical Information with Differentially Private Synthetic Data}

% What is synthetic data?
The idea of synthetic data dates back to \citet{Rubin1993} and \citet{Little1993} who first advanced synthetic data methods in the spirit of multiple imputation. After being formulated as a proper framework \citep{Raghunathan2003}, a series of papers elaborated it further \citep{Abowd2004, Abowd2004a, Reiter2007, Drechsler2010, Kinney2010, Kinney2011}. Different statistical models to generate synthetic data exist, including also machine learning techniques \citep{Reiter2005, Caiola2010, Drechsler2011}.\footnote{For overviews, see also e.g. \citep{Bowen2019, Drechsler2011b, Manrique-Vallier2018, Snoke2016}.}

% Desirable properties of Synthetic data
Synthetic data has a number of desirable properties \citep{Caiola2010}. Synthetic data ideally preserves all relationships between variables. It should be possible to handle diverse data types (categorical, ordinal, continuous, skewed etc.). Computation needs to be feasible also for large data sets. Anybody generating synthetic data should have little work to do to tune the necessary algorithms. If done well, synthetic data has several important (theoretical) advantages over traditional statistical disclosure limitation methods such as data swapping, aggregation or cell suppression. The results are similar no matter whether an analysis is performed on the synthetic or original data. Also, theoretically, synthetic data can protect privacy more easily, because a synthetic data record is not any respondents' actual data record. Finally, data users can draw valid inferences for a variety of estimands -- without having to make any particular assumptions about how the synthetic data were created.

% What is DP
Differential Privacy is a principled privacy framework that helps mitigate the trade-off between sharing statistical information while protecting privacy \citep{differential-privacy, Dwork2013, Nissim2017}. It offers mathematically proven privacy guarantees while allowing to draw inferences about values in the population \citep{Dwork2017}. The key idea is to think about data protection from a relative perspective. What is the relative risk of disclosing information about an individual when running an analysis on a data base that includes their information vs. running an analysis on a data base that does not include their information? It is possible to quantify this privacy difference. Differential Privacy formulates upper bounds about the risk of disclosing information when querying a database in the presence or absence of any one observation. This does not mean that nothing can be learned about an individual: It rather means that if something can be learned about an individual, it does not matter whether an individual is in the data or not \citep{Dwork2010}.

%  DP Synthetic Data
Combining synthetic data with Differential Privacy offers an appealing solution to a number of core challenges: Differentially private synthetic data allow to share statistical information while at the same time protecting privacy in a principled and future proof way. Synthetic data also offers a solution to the query-based limitations of the original Differential Privacy framework. Since Differential Privacy is perserved under post-processing, generating differentially private synthetic data spends the privacy budget only once. In result, any analysis that is run on differentially private synthetic data is also differentially private \citep{Dwork2013, Nissim2017}.

% How to do it
While several methods to generate differentially private synthetic data exist \citep{huang2019,Zhang2017}, the advent of Generative Adversarial Nets (GANs)\footnote{A brief introduction to GANs can be found in the Appendix.} paved the way for general purpose differentially private synthetic data. At first, GANs have been used to address privacy concerns more broadly, for example to protect privacy in images \citep[e.g.][]{Tripathy2017, Wu2018} or -- in contrast -- to attack privacy in the context of distributed learning setting \citep{Hitaj2017}. What ultimately makes GANs differentially private is the injection of the right amount of noise into the training process of the discriminator, since the generator only post-processes the output of the discriminator. \citet{Abadi2016} introduced differentially private stochastic gradient descent, laying an important foundation for later applications to GANs.% \iffalse They clip gradients and add Gaussian noise at each discriminator update step, and then calculate the overall privacy loss $(\epsilon, \delta)$ using so-called moments accounting.\fi
A number of studies follow in their footsteps. \citet{Beaulieu-Jones2017} generate differentially private synthetic data of a medical trial. \citet{Xie2018} also produce differentially private medical data: they too ensure Differential Privacy by combining noise addition and weight clipping. Finally, \citet{Xu2019} apply a combination of adding noise and gradient pruning, and use the Wasserstein distance as an approximation of the distance between probability distributions.

% Balancing Privacy and Data Quality
Private synthetic data can only become a widespread means of disclosure control if it protects privacy while also sharing statistical information. The trade-off comes with a straightforward intuition: Adding privacy means adding noise -- and the noise may affect the statistical information in the data. Even though data privacy is at the heart of research on statistical disclosure limitation, the utility of the resulting data has been a recurrent concern for the discipline in general \citep{Duncan01, purdam2007, trottini2002, taylor2018, Woo2009} and for synthetic data as a means for privacy protection in particular \citep{drechsler2009, Drechsler2011b, Drechsler2018b,reiter2005b,reiter2005c}.

Finding the ``correct'' settings in this trade-off is an optimisation problem. Some studies explicitly express the optimum as an equilibrium in a game-theoretic model \citep[e.g.][]{Chen19, loukides2012, shokri2016}. Others consider the equilibrium between both to be more problematic. From a social choice perspective, privacy and data quality are public goods with actors who may have heterogenous preferences. A data curator -- such as a statistical agency -- can in practice set the optimal levels for privacy and quality \citep{abowd2019}. However, private owners of large confidential data sets may have preferences that lead to the suboptimal provision of these public goods and thus overall welfare \citep{abowd2019b, ghosh2015}.

In the context of private synthetic data, introducing Differential Privacy as a principled data protection framework has certainly been an important step forward. To advance the development of algorithms that are capable of delivering really useful private synthetic data, it is necessary to consider both sides of the coin at the same time -- which puts quality explicitly at eye level with privacy protection. What is data quality? And how could we measure it? A better understanding of the concept and a general framework would certainly be ``extremely beneficial''\citep[][14]{elliot2018}.

%==============================================================================================================
\section{Making Differentially Private Synthetic Data Useful}
Differentially private GANs (DP-GANs) were the first class of models to synthesize data sets from different domains and for different uses. In existing studies, quality is either assessed by looking at the visual quality of generated images \citep[e.g.][]{Xie2018, FrigerioOGD19, TKP20} or the classification evaluation of a natural classification/prediction task \citep[e.g.][]{Beaulieu-Jones2017}. Yet, there is no clear definition of what data quality actually implies beyond these measures. In the absence of natural quality measures it is more difficult to assess the overall quality of synthetic data.

% \iffalse
% \cite{medical, rmsprop_DPGAN, FrigerioOGD19, TKP20}
% We now know how to guarantee privacy in synthetic data. How can we make sure differentially private synthetic data is also useful? Originally, differentially private GANs (DP-GANs) have been built for visual data. It is often enough to take a look at the resulting synthetic images to benchmark the results from different data generating algorithms -- even though images are a very complex distribution to learn. Obviously, researchers from applied fields are much less keen on generating differentially private images. Although image data and ``spreadsheet data'' both come in matrices, their numbers differ in many regards. At the same time, the purpose of the data is different. Both matter for applied researcher's ultimate goal, inference and prediction.
% \fi

\begin{figure}[!ht]
         \centering
         \includegraphics[width=0.9\columnwidth]{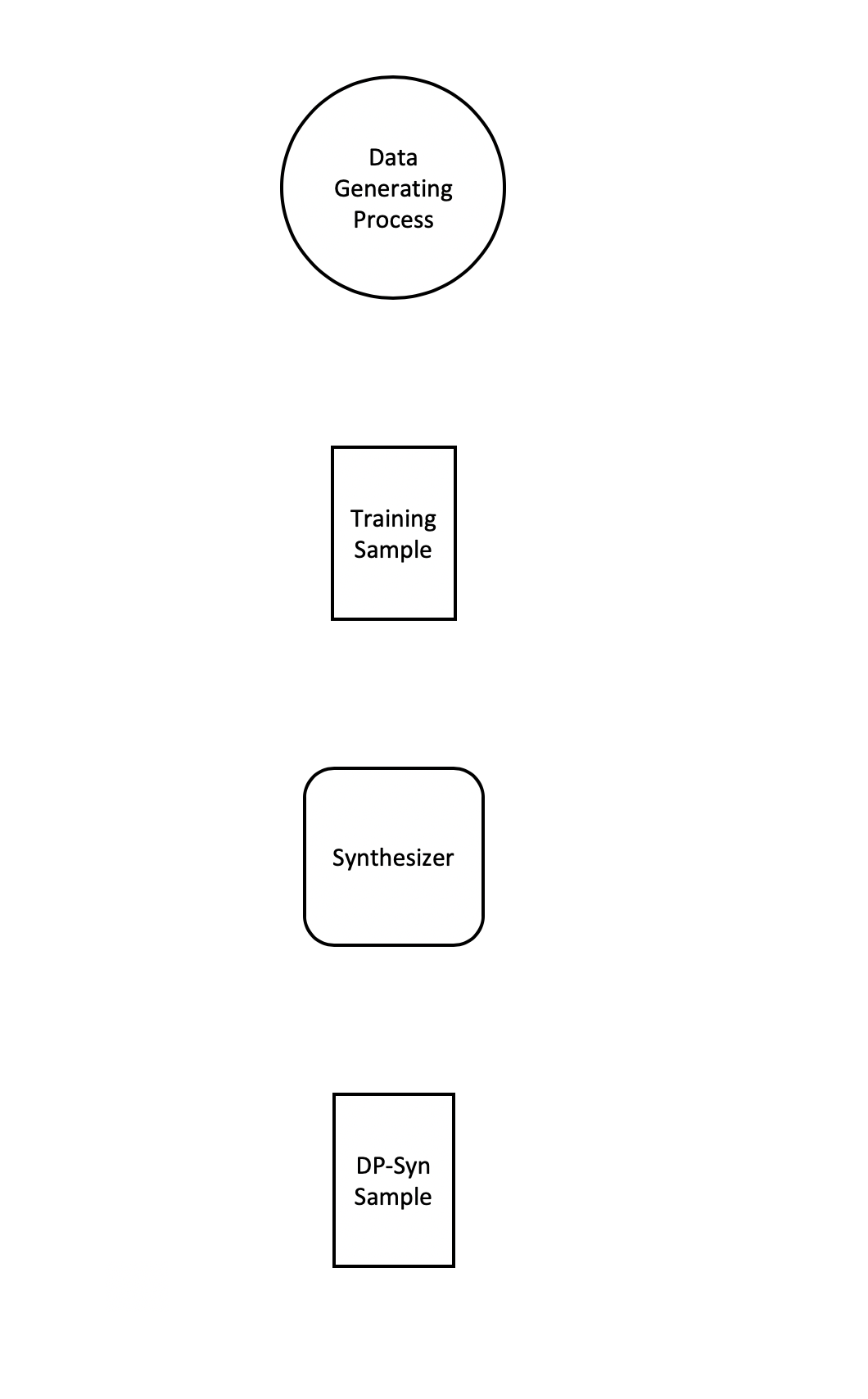}
         \caption{From the Data Generating Process to Differentially Private Synthetic Data.}
         \label{fig:training_data_similarity_2}
\end{figure}

% User Scenario:
% TODO still embedd here
We develop our framework for data quality along the lines of the requirements of a typical data use case: A well intended data curator collects a random sample from a data generating process provided by nature. They have access to this original sample and they would like to share the statistical information. While they might have an idea about the models potential data analysts would typically run, they are ignorant about the exact nature of these analyses. Figure \ref{fig:training_data_similarity_2} shows the different elements along the way from the underlying data generating process to the differentially private synthetic data a curator might want to share with other analysts.

The data curator is uncertain about the intention of the data analysts: are they friendly and well intended data users curious to understand statistical relationships in the population? Or do they have bad intentions and are they rather eager to unveil and exploit information about individuals? Both types of data users have an interest in high data quality -- however, only the bad type would misuse the information. When releasing differentially private synthetic information, the data curator therefore faces an optimization problem: they want to guarantee the highest level of data quality under the constraint of an appropriate level of privacy protection.

\subsection{What Do Applied Researchers Expect from Private Synthetic Data?}
% Inference

Applied researchers may have different expectations when they analyse private synthetic data. Some are interested in rather descriptive analyses of their data. They consider their data not as a sample from what potentially could be a larger population of data sets. This is the case whenever the original data captures a unique data source. For example, studies on political elites such as candidate studies to elections comprise all available data; or data on banking transactions that help identify money laundering are typically also unique instances. Data users analyse this kind of data with a descriptive goal in mind. They do not assume a larger population behind this data about which they want to draw inferences. All analyses they run on synthetic data should ideally lead to similar results as the analyses using the original data.

Other researchers seek to infer information about a population from data. Typically, they draw a random sample from the population to learn more about a true unknown parameter $\theta$. Using their sample, they can calculate an estimate $\hat{\theta}$. If they were hypothetically able to repeatedly draw from the data generating process, these multiple estimates would vary by $V(\hat{\theta})$. Denote $\hat{V}(\hat{\theta})$ the estimate of this variance on the basis of the one sample a researcher has at hand to qualify the certainty of their measurement.

% Inference for DP synthetic data: The problem
This framework works when a researcher can randomly draw from the population and then share this sample with everybody else. However, in the case of differentially private synthetic data this process looks different. Synthesizing data turns inference into a multi-phased problem \citep{Xie2017}: The researcher now has to infer the true underlying population parameter $\theta$ on the basis of the synthesised and private data which was itself generated from a random sample from the population. The researcher has to account for this in their estimate $\hat{\theta}$. In addition, the uncertainty no longer stems from sampling error alone, but it has to account for the synthetization in the estimate of $\hat{V}(\hat{\theta})$, too.
% inference for synthetic data

Conceptually, these two sources of uncertainty are distinct. Data synthetization is very much like multiple imputation -- it was indeed even conceived as a special case of multiple imputation at first \citep{Rubin1993}. Following this line of thinking, \citet{Raghunathan2003} suggest to treat synthetic data as an extreme case of imputation -- basically a data set where all of the values for some unobserved quantities are missing. They propose to generate $m$ synthetic data sets instead of just $1$. Extending the imputation framework then allows to combine the results on the $m$ data sets to estimate $\hat{\theta}$ and $\hat{V}(\hat{\theta})$: The parameter of interest $\theta$ can be estimated as the average $\bar{\theta}$ of all individual estimates $\hat{\theta}_m$. The variance $\hat{V}(\hat{\theta})$ is estimated as a weighted combination of the within variance and between variance of the $m$ data sets.

% Making it DP is a separate step in theory. Not in practice
Valid statistical inference on private synthetic data is possible if the data synthesization process is unbiased and consistent. This is typically the case when the model for synthesization and the model for subsequent analyses are congenial \citep{Meng1994}. Inference about the unknown population parameter $\theta$ on the basis of a synthesized data set $x^{*}$ can then occur in a way that is conceptually similar to the inference framework \citep{Charest2010, Karwa2017}. \citet{Liu2016} shows how to do inference about the unknown population parameter $\theta$ with differentially private synthetic data.\footnote{Her article also extends to the case where both, data synthetization and data sanitisation are not only two conceptually, but actually two analytically different steps \citep{Liu2016}.} It requires an important condition: the synthetization needs to be statistically consistent. This is not a trivial problem. For example, the joint noise process does not only have to be unbiased and ideally also efficient. In addition, any amount of noise that is simply ``sprinkled'' directly on top of the data would bias statistical estimators for nonlinear functions of the data \citep{Evans2019} -- hence the noise needs to enter during data synthetization.

Denote a single data set drawn from the unknown population that is used to train the differentially private synthesizer as $x$. The resulting synthetic data set is $x^{*}$. Assume the model that allows to obtain inference on $\theta$ from the original data $x$ is the same as the model that is used to infer $\theta^{*}$ based on $x^{*}$. According to \citet{Liu2016}, if the estimator that is calculated on $x^{*}$ is consistent for $\theta^{*}$ and the synthetization is also statistically consistent, then $\hat\theta^{*}$ calculated on the synthetic data set is a consistent estimate of the unknown population parameter $\theta$.

\begin{equation}
    \bar{\theta} = \frac{1}{m}\sum\limits_{i=1}^m \hat\theta_i^{*}
\end{equation}

Similar to multiple imputation, generating more than one synthetic data set allows to quantify the uncertainty about the estimate for $\theta$ -- in the light of both, sampling uncertainty to generate $x$ and also synthetization uncertainty to produce $x^{*}$. Assume to generate $m$ data sets $\sum\nolimits_{i=1}^m x_i$. The variance of $\theta$ in the light of the data sets $x_i^{*}$ can then be expressed as
\begin{equation}
\label{eq:rubins_rule}
    u = \bar{\omega} + (1 + \frac{1}{m})b
\end{equation}

again in terms of the average within data-set variance
\begin{equation}
    \bar{\omega} = \frac{1}{m} \sum\limits_{i=1}^m V \left( \hat\theta_i^{*} \right)
\end{equation}

and the between data-set variance

% \iffalse
% \begin{equation}
%     b = \frac{1}{m}\sum\limits_{i=1}^m \left( \hat\theta_i^{*} - \bar{\theta} \right)^2 = b_1 + b_2
% \end{equation}

% where $b_1$ is the variance from synthetization and $b_2$ the variance from sanitization -- a result in sync with the imputation framework from \citet{Rubin1993}.
% \fi

\begin{equation}
    b = \frac{1}{m - 1}\sum\limits_{i=1}^m \left( \hat\theta_i^{*} - \bar{\theta} \right)^2
\end{equation}

-- a result in sync with the imputation framework from \citet{Rubin1993}, often cited as Rubin's rules.

In reality, however, the private data synthesizer and the model for analysis are often uncongenial \citep{Meng1994}: While congenial differentially private data synthesizers do exist -- and allow to exactly specify bias and efficiency of the private synthetic data \citep[e.g.][]{McClure2012} -- the reality often looks different. The proposed models are highly specialized and tailored to a specific purpose. Flexible, general use private synthetic data can hardly be released with these models. Congeniality could be achieved if the data curator succeeds in anticipating and hence nesting (all) models of the analyst. Even though non-parametric synthesizers such as CART are known to be capable of covering a broad array of subsequent models for analysis \citep{Reiter2005}, they yet have to be made differentially private \citep{Shlomo2018}.

\citet{Xie2017} propose a general multi-phase inference framework in the face of uncongeniality. It is not enough if each, the imputer and the analyst, use statistically valid models for themselves. Both models need to be congenial to allow for inferring the true unknown data generating process. If they are uncongenial, inference on the basis of the data $x^{*}$ may be unbiased, but it may lead to wrong conclusions about the efficiency of the estimate. Two scenarios are possible: If the imputer's model is more saturated than the analysts model, Rubin's rules still hold and are confidence proper. However, in the second scenario -- if the imputers' model is less saturated than the analyst's model -- Rubin's rules are super efficient and propose too narrow confidence bounds. \citet{Xie2017} show how to be better save than sorry and prove upper uncertainty bounds: Doubling the variance estimate -- $u$ from Rubin's rule in equation \ref{eq:rubins_rule} -- is enough to generate a conservative estimate for multi-phase inference even under uncongenial models. In line with \citet{Xie2017}, we therefore propose to adapt the estimate of the uncertainty

\begin{equation}
    u_{uc} = 2 u
\end{equation}

where $u_{uc}$ expresses the variance under uncongeniality.

Another goal for applied researchers is prediction. If a researcher wants to implement differentially private out-of-sample prediction, the process would look as follows. First, the data curator samples data $x$ from the population and trains the private synthesizer on $x$ to release the synthetic data $x^{*}$. The data analyst then learns the model $m^{*}$ using data $x^{*}$ to predict outcome $y^{*}$. New data $x^{'}$ is coming in which is drawn from the same population as data $x$. The analyst uses this new data $x^{'}$ and the model $m^{*}$ to predict outcome $y$.

%==========================================================================================
\section{A General Framework to Evaluate the Quality of Private Synthetic Data}
% Ramp: No one has had a closer look at data quality for micro data
So where can we download these differentially private synthetic micro data wonder machines? Unfortunately, while first synthesizers have proven themselves in the controlled settings of pilot studies, general models that are both secure and useful still need to improve to be really useful. In addition, little has been done to understand the algorithms' statistical properties. To advance the development of differentially private micro data synthesizers, we propose a joint data challenge. We hope that it serves to coordinate efforts of scholars with backgrounds in various academic disciplines.

\subsection{Proposal for a Joint Benchmark}

% The different challenges
Applied researchers' micro data have a number of characteristics that make it challenging to generate differentially private synthetic data for them. Indeed, data can be continuous. More often, however, micro-data is discrete. Categories could be ordinal or without any particular natural order to them. Structural zeros preclude logically impossible values such as pregnant three-year-olds. Often, parts of the data are missing -– either at random or not at random. Finally, data structures can be nested. In time series data, the same individual might be observed at different points in time. In hierarchical data, individuals are part of a larger group that comes with its own measurements, such as students in a classroom.

% \iffalse
% We define four different scenarios under which differentially private data synthesizers should work well (Table~\ref{tab:datachallenges}). The first scenario is a baseline scenario that captures a realistic setting for micro data. Data has continuous variables, categorical variables and structural zeros.\footnote{We explicitly define a data generating process in the appendix.} The next scenario also includes missing data, both completely at random and also at random. The third scenario defines nested data structures. Finally, scenario number four includes all those settings. So far, algorithms have been developed that are capable of generating differentially private data for the baseline scenario. To our knowledge, there is no technical solution for all others scenarios.
% \fi

% \usepackage{booktabs}
\begin{table*}[!htbp]
\centering
\caption{Typical Data Challenges Applied Data Analysts Are Facing.}
\label{tab:datachallenges_2}
\begin{tabular}{@{}lcccccc@{}}
\toprule
           & \multicolumn{1}{l}{Continuous} & \multicolumn{1}{l}{Discrete} & \multicolumn{1}{l}{Structural}        & \multicolumn{1}{l}{Missing} & \multicolumn{1}{l}{Nested} \\
           & \multicolumn{1}{l}{Data}       & \multicolumn{1}{l}{Data}     & \multicolumn{1}{l}{Zeros}             & \multicolumn{1}{l}{Data}    & \multicolumn{1}{l}{Data}   \\
\midrule
Scenario 1 & \checkmark                     & \checkmark                       & \checkmark                        &                           &  \\
Scenario 2 & \checkmark                     & \checkmark                       & \checkmark                        & \checkmark                &                    \\
Scenario 3 & \checkmark                     & \checkmark                       & \checkmark                        &                         &      \checkmark        \\
Scenario 4 & \checkmark                     & \checkmark                       & \checkmark                        & \checkmark              &      \checkmark        \\
\bottomrule
\end{tabular}
\end{table*}

Data sets typically include a combination of those data characteristics. Since any combination of the characteristics is possible, there are 36 potential scenarios. For illustration purposes we focus on the four different scenarios in Table~\ref{tab:datachallenges_2}. The first scenario is a baseline scenario that captures a realistic setting for micro data. Data has continuous variables, categorical variables and structural zeros.\footnote{We explicitly define a data generating process in the appendix.} The second scenario also includes missing data, both completely at random and also at random. The third scenario defines nested data structures. Finally, scenario number four includes all those settings. So far, algorithms have been developed that are capable of generating differentially private data for the baseline scenario. To our knowledge, there is hardly any research on synthesizers for the other scenarios.

% Different Disciplines
A data synthesizer should be able to rise to the challenges in these scenarios in two different \emph{disciplines}. First, differentially private data synthesizers ideally perform well at different levels of privacy protection. The literature considers values at $\epsilon=0.1$ up to $\epsilon=1$ acceptable. A value of $\epsilon=5$ is already quite a considerable privacy leak, but, depending on the data, still worthwhile exploring. Second, the algorithms need to perform at different sample sizes. Applied researchers' sample sizes range from clinical studies ($500$ individuals) to surveys with $10,000$ individuals. Data sets with more than $100,000$ individuals do exist, but are fairly rare.
% \iffalse\footnote{After all, ``there is essentially only one scientific way to evaluate and compare statistical procedures -- show how they work when applied repeatedly, in reality, via simulation or in thought experiments.\citep{Xie2017}''}\fi

\subsection{Measuring Data Quality}
% TODO shorter on dimension 1
In a perfect world, a differentially private data synthesizer should work well for any of the 36 possible combinations of data challenges and across the different disciplines. We now turn to what we mean by ``working well''.

Data quality is context dependent. Measures for the quality of differentially private synthetic data should reflect the needs of those who work with this kind of data. We suggest to distinguish between quality measures along two dimensions.%\iffalse The first dimension is concerned with the similarity of the differentially private synthetic data $x^{*}$.\fi
The first dimension distinguishes between the similarity of the differentially private synthetic data $x^{*}$ to the training data (i.e. the data that was used to generate the synthetic data) and the generalization properties of $x^{*}$. On the one hand, data users might be particularly concerned with the similarity of $x^{*}$ to the original training data $x$. On the other hand, data users might care less about similarity of the data set to one particular training set. Instead, data analysts may be interested in inferring information about the underlying population -- to either identify relationships between variables or build models for out-of-sample predictions.

The second quality dimension is concerned with general data quality and specific data quality \citep{Snoke2018}\footnote{Note that \citet{Snoke2018} speak of general and specific utility.}. General data quality describes the overall distributional similarity between the real data and synthetic data sets. In turn, to reflect specific analyses data analysts expect to run on synthetic data we introduce specific data quality. This could be the similarity of marginal distributions, the performance of regression models or the predictive performance of machine learning models trained on synthetic data sets as compared to the same analysis carried out on the real data.
% TODO add footnote to point out relative perspective of marginal similarities. Define general and specific utility better and in line with Snoke et al.

% What data utility measures to chose from
% Grammar when explaining: a) what do you want to do b) how does it work c) what is a good and a bad value
Table~\ref{tab:utility_2} provides an overview of data quality measures along both dimensions, resulting in overall four different quadrants. The ultimate choice of the quality measures is indeed subjective and needs to fit their purpose \citep{Shlomo2018}. We first turn to data quality of private synthetic data $x^{*}$ relative to the training data $x$ in the top left cell of Table~\ref{tab:utility_2}.

\begin{table*}[!htbp]
\centering
\caption{Proposed Quality Measures for Private Synthetic Data Along Two Dimensions.}
\label{tab:utility_2}
\begin{tabular}{@{}p{2.3cm}p{5.3cm}p{5.3cm}@{}}
                           & \textbf{General Data Quality}                  & \textbf{Specific Data Quality} \\
\midrule
\textbf{Training Data Similarity} & Similarity of DP synthetic data and sample 1-way marginals: Wasserstein Randomisation Test         & Compare $\hat\theta^{*}$ with $\hat\theta$: average percent bias (difference between coefficients calculated on synthetic data and training data)                \\
\addlinespace
                           & $pMSE$ bewteen synthetic data and training sample                 & Compare $\hat V^{*} \left(\hat\theta^{*}\right)$ with $\hat V \left(\hat\theta\right)$: average variance and covariance ratio \\
% \addlinespace
                        %   &         &  Prediction RMSE w.r.t. sample           \\
\midrule
\textbf{Generalisation Similarity}   &  Similarity of DP synthetic data and independent test sample 1-way marginals: Wasserstein Randomisation Test  & Compare $E\left(\hat\theta^{*}_{m}\right)$ with $\theta$: average percent bias (difference between true population values and coefficients on synthetic data)\\
\addlinespace
                           & $pMSE$ bewteen synthetic data and independent test samples     & Compare $E\left(\hat V^{*}_{m} \left(\hat\theta^{*}_{m}\right)\right)$ with $V\left(\theta\right)$: coverage, average width \\
\addlinespace
                           &  & Prediction RMSE w.r.t. further samples from population             \\

\end{tabular}
\end{table*}

To compare the overall similarity of the joint distribution and to calculate the general quality of the synthetic copies we rely on the propensity score mean squared error (pMSE) ratio score for synthetic data as proposed by \citet{Snoke2018}. The pMSE requires training a discriminator (e.g. a CART model) that is capable to distinguish between real and synthetic examples. A synthetic data set has high general data quality, if the model cannot distinguish between real and fake examples. It is possible to approximate the pMSE by resampling the null distribution of the pMSE. The pMSE ratio score divides the pMSE by the expectation of the null distribution. A pMSE ratio score of 0 would indicate that the synthetic and real data are identical.
% \iffalse Next, to compare the overall similarity of the joint distribution, we rely on the propensity-score mean-squared error ($pMSE$) \citet{Snoke2018}. Their procedure first match data from the original and the training set and then empirically investigate to which extent powerful prediction models (e.g. CART) are capable of distinguishing whether an observation originates from the training data set or the synthetic data set. A ratio $pMSE = 1$ indicates correct synthesis of the data. \fi
% Eventually: $pMSE$ longer. Or Wasserstein Distance
One of the most relevant concerns when releasing differentially private synthetic data remains the similarity of $1-way$ marginals. The Wasserstein distances measures the difference between two univariate distributions with unknown statistical properties \citep{Robinson2017}. To compare this data specific metric across different attributes, it has to be put into relative perspective: We propose a metric based on the idea of randomization inference \citep{Basu1980, Fisher1935}. Each data point is randomly assigned to one of two data sets and the similarity of the resulting two distributions is measured with the Wasserstein distance. Repeating this random assignment a great number of times (e.g. $10'000$ times) provides an empirical approximation of the distances' null distribution. Similar to the pMSE ratio score we then calculate the ratio of the measured Wasserstein distance and the median of the null distribution to get a Wasserstein distance ratio score that is comparable across different attributes. Again a Wasserstein distance ratio score of 0 would indicate that two marginal distributions are identical. Larger scores indicate greater differences between distributions. %\iffalse\footnote{Please note the similarity to randomization inference \citep{Basu1980, Fisher1935}.}  How often is the Wasserstein distance between the original data and the synthetic data larger than between two randomly assigned data sets? The resulting proportion can be interpreted as a $p$-value, and we suggest to follow the convention. If $p > 0.05$, then the two marginal distributions are very close and the data quality is high. Instead, if $p < 0.05$, both marginals are fairly distinct, and hence data quality is low. To measure the overall similarity between original data and synthetic data in its margins, we calculate the proportion of Wasserstein distances larger than the p-value $pWD_{1d}$ for all $1-way$ marginals, and $pWD_{2d}$ for all $2-way$ marginals.\fi

The similarity between the private synthetic data and the training data addresses data analysts' specific requirements for data quality.\footnote{For similar measures used in multiple imputation, see \citet{vanBuuren2018}.} They might first be interested in understanding correlations -- and multivariate statistical models quantify these correlations while controlling for the influence of other variables. It is straightforward to calculate bias of parameters $\hat \theta^{*}$ estimated on synthetic data $x^{*}$ in comparison to $\hat \theta$ estimated on $x$ as percent bias with their normalised difference. We then aggregate all biases to one score by taking their average. To compare the variances, we follow \citet{Drechsler2018}, and measure the ratio between the variances of the parameters $\hat V^{*} \left(\hat\theta^{*}\right)$ in the private synthetic data and divide it with the variances $\hat V \left(\hat\theta\right)$ of the original training data. The resulting variance ratio equals 1 if they are both the same. Averaging across all ratios expresses their general difference. % Finally, the RMSE allows to compare the predictive performance of the models $m^{*}$ learned on the data $x^{*}$. Simply use the model $m^{*}$ to predict outcomes using data from the unseen original data $x$ -- and tell the difference using the RMSE.

\begin{figure}[!ht]
         \centering
         \includegraphics[width=0.9\columnwidth]{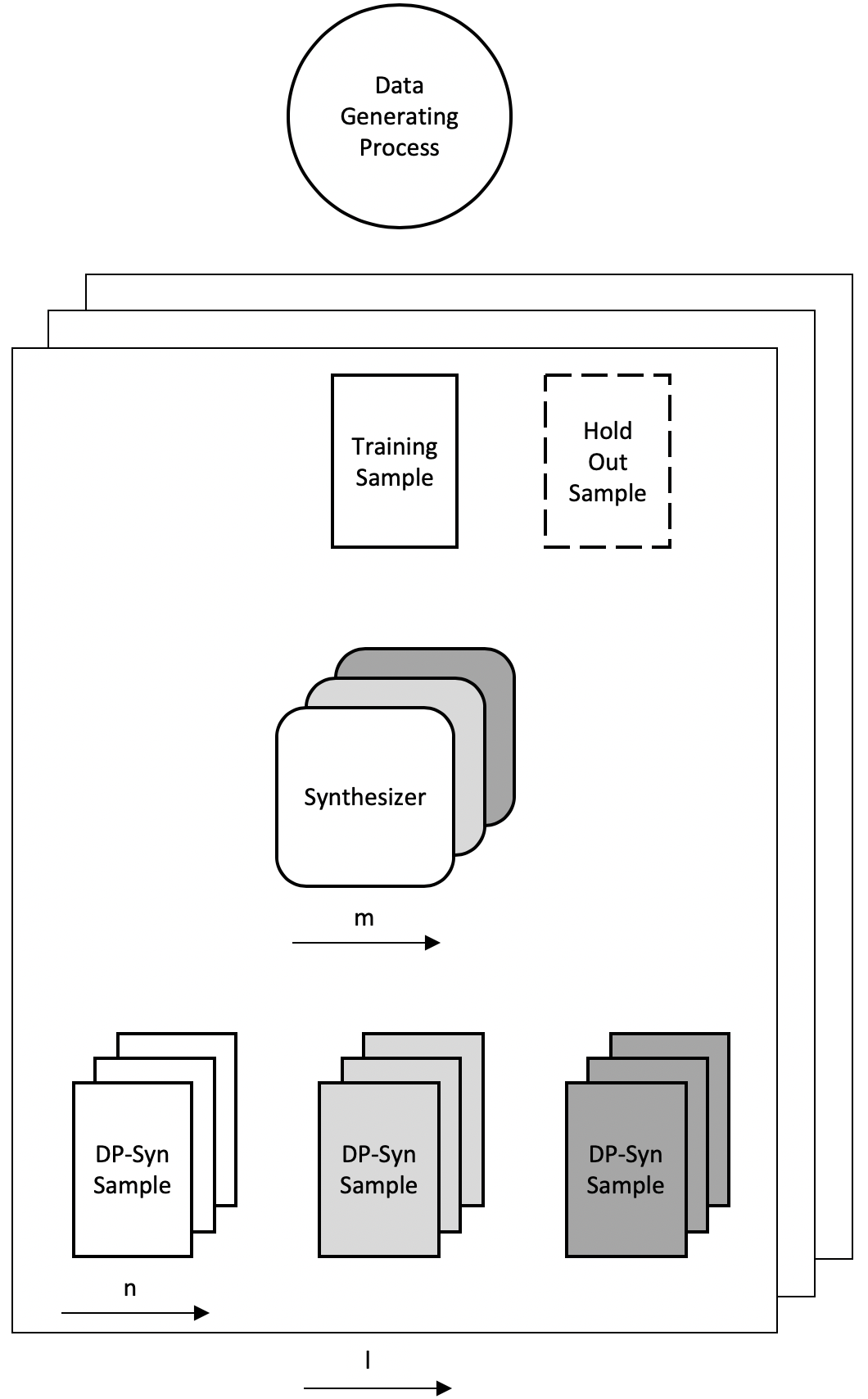}
         \caption{The Setup of the Benchmark.}
         \label{fig:benchmark_2}
\end{figure}

In other cases, analysts are more interested in using samples to generalise beyond the observations in the training set and make inferences about a larger population. When using private synthetic data in this setting, the data analyst needs to account for three layers of random errors in this data, first sampling error when drawing the training data from the underlying population, second model uncertainty when training the data synthesizer (even without Differential Privacy) and third the privacy noise that is introduced to the training of the data synthesizer to release differentially private synthetic data. We suggest to empirically validate data quality in a Monte Carlo framework and average results across $1'000$ differentially private synthetic data sets.\footnote{We chose the sample size for pragmatic reasons. Some of the differentially private data synthesizers may take non-negligible time for training. In the light of today's available computational power we acknowledge that $1'000$ data sets may be not enough to complete integrate over all uncertainty from the three respective steps. We chose $10^3$ data sets in line with what \citet{Rubin1996} considered sufficient for imputation.} They are the result of $l = 10$ different training data sets -– with another $10$ independent data sets held out as test sets. Then, train $m = 10$ independent data synthesizers on each of the training sets and finally release $n = 10$ data sets from each of these synthesizers. Figure \ref{fig:benchmark_2} shows the setup for the benchmark.

The first relevant measure for general data quality is the similarity of marginals, again calculated with the Wasserstein distance ratio score. This time, however, we pair each private synthetic data set with the respective held out data sets to understand the capacity to generalize. %\iffalse We calculate the overall similarity of the synthetic data with the held out data and count the overall percentage of data sets that fall below a similarity threshold of $p < 0.05$. As before, there is one measure for all 1-way marginals and another overall measure for $2-way$ marginals.\fi
In a similar vein, the $pMSE$ ratio score again summarizes the overall similarity between joint distribution. We calculate it against the hold out test sets and summarize by averaging across all $pMSE$ ratios.

For the specific data quality measures, the analyst might be concerned with bias of the parameters $\hat \theta^{*}_{i}$ of a multivariate regression model. We calculate bias for each of the synthetic data sets, comparing it to the values that define the true underlying data generating process. Standardising and then averaging across all parameters gives an overall percentage value that summarises bias in general. The coverage rate of the variance captures efficiency \citep{Drechsler2018}. How often does the 90\% confidence interval include the true value from the population? Averaging across all samples and cases yields one single measure. Finally, an analyst might be interested in generating a model that is capable of predicting values for out-of-sample prediction. Here, we estimate a model on each of the synthetic datasets. We then predict values for unseen data in the held out validation sets and summarise all results with an average RMSE.

% \iffalse
% Figure \ref{fig:analysisplan1} provides two charts that summarise the procedure for training data similarity (Figure~\ref{fig:training_data_similarity}) and generalisation similarity (Figure~\ref{fig:generalisation_similarity}).

% \fi

% \begin{figure}
%     \centering
%     \includegraphics[width = .35\textwidth]{figures/data_challenge.png}
%     \caption{}
%     \label{fig:analysisplan1}
% \end{figure}

\subsection{The Data Challenge}
\begin{figure}
     \centering

         \includegraphics[width=0.8\columnwidth]{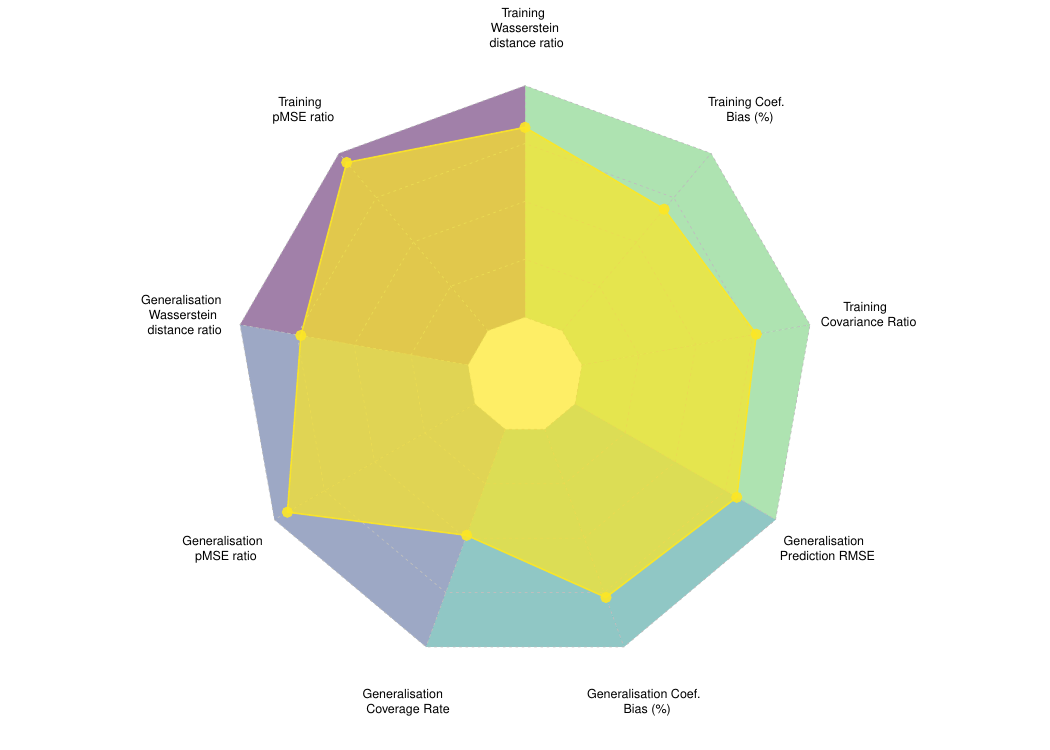}
         \caption{Summary Chart for Data Quality.}
         \label{fig:summary_2}

\end{figure}

% Doing it
We envision a data challenge that tests differentially private data synthesizers for potentially different data scenarios (e.g. the baseline scenario, missing data, nested data) in the two related disciplines (training data size and privacy budget). For the challenge, we measure data quality according to Table \ref{tab:utility_2}. For a quick comparison of different differentially private data synthesizers we summarize the measures in an easy to read chart (see Figure \ref{fig:summary_2}). In this hypothetical example, the yellow polygon displays the performance of a differentially private data synthesizer for the nine measures from above. The further out (the bigger) the edges of the polygon, the better the performance of a synthesizer in that domain. With this chart it is also easy to see whether a particular data synthesizer does well for some measures of data quality, while it fails for others. The ultimate goal should be to balance the performance for all measures.

To document the progress of differentially private data synthesizers, we are maintaining an internet repository at \url{privatesyntheticdata.github.io}. We also use the homepage to describe the data challenge more fully. You can learn more about the details for replication, i.e. a summary of the exact specifications for each scenario and each of the three ``disciplines''. We understand that researchers who submit a new algorithm do not undermine the challenge, e.g. by tailor suiting the algorithm to the particular \emph{ex-ante} known data generating process. We also trust that any hyperparameter optimization occurs without explicitly overfitting in a non-private way.

\section{Results for a DP-GAN}

We now empirically evaluate an easy to implement DP-GAN to establish a first score on our proposed benchmark. Training data is drawn from a data generating process that is in line with our Scenario 1 in Table \ref{tab:datachallenges}.\footnote{Details on the data generating process can be found in the Appendix.} % \iffalse from a Social Scientist's perspective, focusing in particular on the trade-off between data utility and privacy protection. We turn to a simulation study and evaluate the usability of the synthetic data generated by a DP-GAN with respect to a measure of general utility for synthetic data \citep{Snoke2018}. Furthermore, we check out correlation structures and regression coefficients in resulting synthetic data at various levels of $\epsilon$.\fi
Our training sample has $10'000$ observations and 8 attributes (four attributes with continuous data, four attributes with discrete data, one of which contains structural zeros). The GAN network consists of three fully connected hidden layers (256, 128 and 128 neurons) with Leaky ReLu activation functions. We add dropout layers to the hidden layers of the generator with a dropout rate of 50\%. The latent noise vector $Z$ is of dimension 32 and independently sampled from a gaussian distribution with mean 0 and standard deviation of 1. To sample from discrete attributes we apply the Gumbel-Softmax conversion \citep{maddison2016concrete, jang2016categorical} to the output layer of the Generator. During training of the DP-GAN we update the weights of the Discriminator with the DP Adam optimizer as implemented in \texttt{tensorflow\_privacy}. We keep track of the values of $\epsilon$ and $\delta$ by using the moments accountant \citep{Abadi2016, Mironov17}. Our final value of $\epsilon$ is $1$ and $\delta$ is $\frac{1}{2N} = 5\times10^{-5}$, after training the DP-GAN for $2,000$ update steps with mini-batches of 100 examples randomly chosen from the training data.

To evaluate this DP-GAN with our benchmark we produce $l = 10$ training data sets and we train $m = 10$ of our DP-GANs on each of the training data sets. Finally, we produce $n = 10$ differentially private synthetic data sets from each of the trained DP-GANs, leaving us in total with $1'000$ differentially private synthetic data sets to assess the data quality.

\begin{figure}[!ht]
     \centering

         \includegraphics[width=0.8\columnwidth]{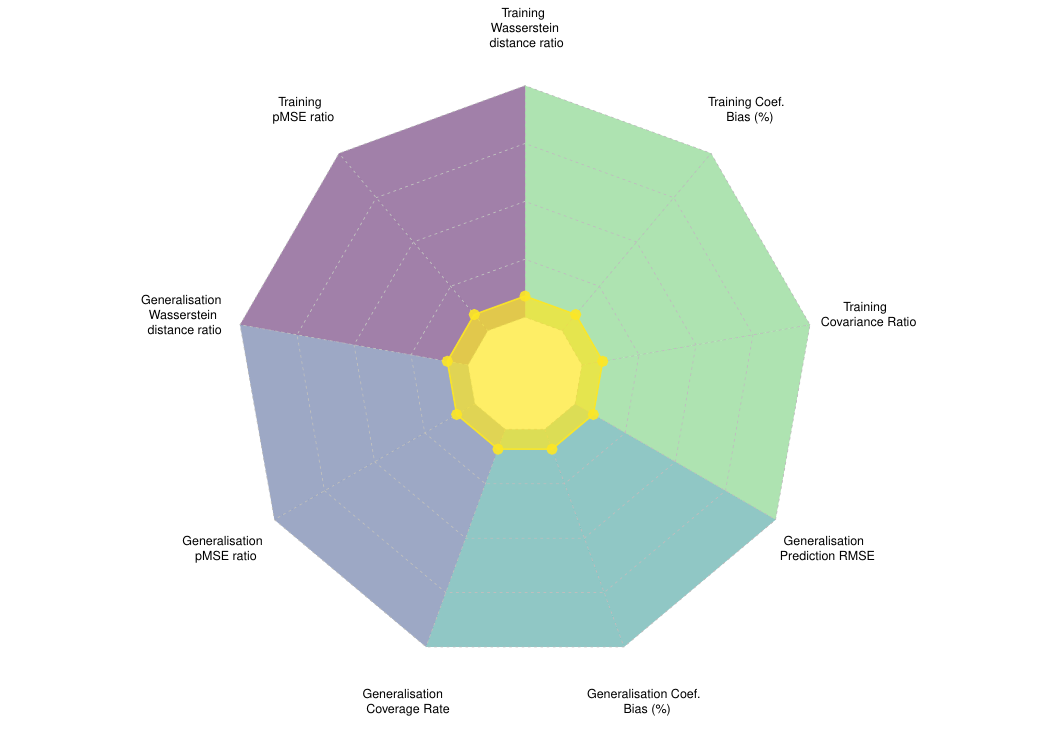}
         \caption{Summary Chart for our baseline DP-GAN.}
         \label{fig:results_2}

\end{figure}

Figure \ref{fig:results_2} shows the performance of our DP-GAN for the nine quality measures. Since we do not expect our DP-GAN to achieve a state of the art performance and rather think of it as a first simple baseline, we initialize the summary chart such that our DP-GAN is very close to the center\footnote{The best score for each of the nine measures is 0, to define the worst score we take our realized score and add 10\%.}. Table \ref{tab:results_2} shows the individual values of the nine scores.

\begin{table}[!ht]
\centering
\caption{Results for our baseline DP-GAN.}
\label{tab:results_2}
\begin{tabular}{l|r}
\toprule
Training
 Wasserstein
 distance ratio & 10.09\\

Training
 pMSE ratio & 3522.96\\

Generalisation
 Wasserstein
 distance ratio & 10.22\\

Generalisation
 pMSE ratio & 3505.01\\

Generalisation
 Coverage Rate & 0.89\\

Generalisation Coef.
  Bias (\%) & 119.93\\

Generalisation
 Prediction RMSE & 9.39\\

Training
 Covariance Ratio & 7.72\\

Training Coef.
 Bias (\%) & 115.52\\
\bottomrule
\end{tabular}
\end{table}

Without comparing these scores to other differentially private data synthesizers interpretation of the scores is hard. Therefore, we think of our first scores as the baseline, leaving other approaches plenty of room for. improvement.

\section{Conclusion}

% \iffalse What does it mean if an algorithm is good in some fields but not in other? %TODO
% Not long ago, \citet{McClure2012} noted that ``[t]here is a long way to go before differentially private synthetic data generation becomes feasible for highly complex datasets.'' In this case, the future came much faster than expected. Recent advances in Computer Science and Statistics allow to generate highly convincing differentially private synthetic visual and audio data. However, the same algorithms are currently of limited use when generating micro data applied researchers are typically working with.\fi

Building on scenarios that describe applied researchers' main data related obstacles we seek to contribute to the development of differentially private data synthesizers and propose a joint and unified benchmark. It is our hope that by formulating a problem that needs to be solved, we offer a focus point for further collaboration across disciplines. Our suite of well defined data generating processes illustrates typical challenges of applied researchers' data, i.e. discrete data, structural zeros, missing data and nested data. Data generating algorithms should be able to account for these challenges in two different settings: at different levels of privacy and at different amounts of available training data. Finally, to assess the quality of the resulting synthetic data from the vantage point of applied researchers we introduce a set of appropriate data quality measurements along two dimensions: On the one hand similarity towards the training data or the underlying population, on the other hand general and specific data quality. All differentially private data synthesizers should ideally be capable of rising to these challenges under different settings: at different levels of privacy and with different training data sizes. They ideally also synthesize data that is similar to the training sample and generalizes well to the underlying population.
Differentially private synthetic data contributes to solving important current challenges. It offers principled privacy guarantees, opening the door to sharing data freely. Access to the statistical information in all the data currently collected promises to open the door wide to a great range of positive externalities -- be they in science, government or business.

\newpage

\bibliography{references}

\begin{thebibliography}{70}
\providecommand{\natexlab}[1]{#1}
\providecommand{\url}[1]{\texttt{#1}}
\expandafter\ifx\csname urlstyle\endcsname\relax
  \providecommand{\doi}[1]{doi: #1}\else
  \providecommand{\doi}{doi: \begingroup \urlstyle{rm}\Url}\fi

\bibitem[Abadi et~al.(2016)Abadi, Chu, Goodfellow, McMahan, Mironov, Talwar,
  and Zhang]{Abadi2016}
Abadi, M., Chu, A., Goodfellow, I., McMahan, H.~B., Mironov, I., Talwar, K.,
  and Zhang, L.
\newblock Deep learning with differential privacy.
\newblock \emph{Proceedings of the 2016 ACM SIGSAC Conference on Computer and
  Communications Security}, Oct 2016.
\newblock \doi{10.1145/2976749.2978318}.
\newblock URL \url{http://dx.doi.org/10.1145/2976749.2978318}.

\bibitem[Abowd \& Lane(2004)Abowd and Lane]{Abowd2004}
Abowd, J.~M. and Lane, J.
\newblock {New Approaches to Confidentiality Protection: Synthetic Data, Remote
  Access and Research Data Centers}.
\newblock In Domingo-Ferrer, J. and Torra, V. (eds.), \emph{Privacy in
  Statistical Databases}, volume 3050, pp.\  282--289. 2004.
\newblock ISBN 978-3-540-22118-0.
\newblock \doi{10.1007/978-3-540-25955-8_22}.

\bibitem[Abowd \& Schmutte(2019)Abowd and Schmutte]{abowd2019}
Abowd, J.~M. and Schmutte, I.~M.
\newblock An economic analysis of privacy protection and statistical accuracy
  as social choices.
\newblock \emph{American Economic Review}, 109\penalty0 (1):\penalty0 171--202,
  January 2019.
\newblock \doi{10.1257/aer.20170627}.
\newblock URL \url{https://www.aeaweb.org/articles?id=10.1257/aer.20170627}.

\bibitem[Abowd \& Woodcock(2004)Abowd and Woodcock]{Abowd2004a}
Abowd, J.~M. and Woodcock, S.~D.
\newblock Multiply-imputing confidential characteristics and file links in
  longitudinal linked data.
\newblock In Domingo-Ferrer, J. and Torra, V. (eds.), \emph{Privacy in
  Statistical Databases}, pp.\  290--297, Berlin, Heidelberg, 2004. Springer
  Berlin Heidelberg.
\newblock ISBN 978-3-540-25955-8.
\newblock \doi{10.1007/978-3-540-25955-8_23}.

\bibitem[Abowd et~al.(2019{\natexlab{a}})Abowd, Schmutte, Sexton, and
  Vilhuber]{abowd2019b}
Abowd, J.~M., Schmutte, I.~M., Sexton, W., and Vilhuber, L.
\newblock Suboptimal provision of privacy and statistical accuracy when they
  are public goods, 2019{\natexlab{a}}.

\bibitem[Abowd et~al.(2019{\natexlab{b}})Abowd, Schmutte, Sexton, and
  Vilhuber]{abowd2019c}
Abowd, J.~M., Schmutte, I.~M., Sexton, W.~N., and Vilhuber, L.
\newblock Why the economics profession must actively participate in the privacy
  protection debate.
\newblock \emph{AEA Papers and Proceedings}, 109:\penalty0 397--402, May
  2019{\natexlab{b}}.
\newblock \doi{10.1257/pandp.20191106}.
\newblock URL \url{https://www.aeaweb.org/articles?id=10.1257/pandp.20191106}.

\bibitem[Basu(1980)]{Basu1980}
Basu, D.
\newblock Randomization analysis of experimental data: The fisher randomization
  test.
\newblock \emph{Journal of the American Statistical Association}, 75\penalty0
  (371):\penalty0 575--582, 1980.
\newblock ISSN 01621459.
\newblock \doi{10.2307/2287648}.
\newblock URL \url{http://www.jstor.org/stable/2287648}.

\bibitem[Beaulieu-Jones et~al.(2019)Beaulieu-Jones, Wu, Williams, Lee,
  Bhavnani, Byrd, and Greene]{Beaulieu-Jones2017}
Beaulieu-Jones, B.~K., Wu, Z.~S., Williams, C., Lee, R., Bhavnani, S.~P., Byrd,
  J.~B., and Greene, C.~S.
\newblock Privacy-preserving generative deep neural networks support clinical
  data sharing.
\newblock \emph{Circulation: Cardiovascular Quality and Outcomes}, 12\penalty0
  (7):\penalty0 e005122, 2019.
\newblock \doi{10.1161/CIRCOUTCOMES.118.005122}.

\bibitem[Benschop et~al.(2019)Benschop, Machingauta, and Welch]{Benschop2019}
Benschop, T., Machingauta, C., and Welch, M.
\newblock Statistical disclosure control: A practice guide.
\newblock 2019.

\bibitem[Bowen \& Liu(2020)Bowen and Liu]{Bowen2019}
Bowen, C.~M. and Liu, F.
\newblock Comparative study of differentially private data synthesis methods.
\newblock \emph{Statist. Sci.}, 35\penalty0 (2):\penalty0 280--307, 05 2020.
\newblock \doi{10.1214/19-STS742}.
\newblock URL \url{https://doi.org/10.1214/19-STS742}.

\bibitem[Caiola \& Reiter(2010)Caiola and Reiter]{Caiola2010}
Caiola, G. and Reiter, J.~P.
\newblock Random forests for generating partially synthetic, categorical data.
\newblock \emph{Trans. Data Privacy}, 3\penalty0 (1):\penalty0 27--42, apr
  2010.
\newblock ISSN 1888-5063.
\newblock \doi{10.5555/1747335.1747337}.

\bibitem[Charest(2010)]{Charest2010}
Charest, A.-S.
\newblock How can we analyze differentially-private synthetic datasets?
\newblock \emph{Journal of Privacy and Confidentiality}, 2\penalty0
  (2):\penalty0 21--33, 2010.
\newblock \doi{10.29012/jpc.v2i2.589}.

\bibitem[Chen et~al.(2019)Chen, Taub, and Elliot]{Chen19}
Chen, Y., Taub, J., and Elliot, M.
\newblock {Trade-off between Information Utility and Disclosure Risk in GA
  Synthetic Data Generator}.
\newblock In \emph{{Joint UNCECE/EUROSTAT Work Session on Statistical Data
  Confidentiality}}, 2019.
\newblock URL
  \url{http://www.unece.org/fileadmin/DAM/stats/documents/ece/ces/ge.46/2019/mtg1/SDC2019_S3_UK_Chen_Taub_Elliot_AD.pdf}.

\bibitem[Drechsler(2011)]{Drechsler2011b}
Drechsler, J.
\newblock \emph{{Synthetic Datasets for Statistical Disclosure Control}}.
\newblock Springer, 2011.
\newblock \doi{10.1007/978-1-4614-0326-5}.

\bibitem[Drechsler(2018)]{Drechsler2018}
Drechsler, J.
\newblock Some clarifications regarding fully synthetic data.
\newblock In Domingo-Ferrer, J. and Montes, F. (eds.), \emph{Privacy in
  Statistical Databases}, pp.\  109--121, Cham, 2018. Springer International
  Publishing.
\newblock ISBN 978-3-319-99771-1.
\newblock \doi{10.1007/978-3-319-99771-1_8}.

\bibitem[Drechsler \& Reiter(2009)Drechsler and Reiter]{drechsler2009}
Drechsler, J. and Reiter, J.
\newblock {Disclosure Risk and Data Utility for Partially Synthetic Data: An
  Empirical Study Using the German IAB Establishment Survey}.
\newblock \emph{Journal of Official Statistics}, 25\penalty0 (4):\penalty0
  589--603, 2009.

\bibitem[Drechsler \& Reiter(2010)Drechsler and Reiter]{Drechsler2010}
Drechsler, J. and Reiter, J.
\newblock {Sampling With Synthesis: A New Approach for Releasing Public Use
  Census Microdata}.
\newblock \emph{Journal of the American Statistical Association}, 105\penalty0
  (492):\penalty0 1347--1357, 2010.
\newblock \doi{10.1198/jasa.2010.ap09480}.

\bibitem[Drechsler \& Reiter(2011)Drechsler and Reiter]{Drechsler2011}
Drechsler, J. and Reiter, J.
\newblock An empirical evaluation of easily implemented, nonparametric methods
  for generating synthetic datasets.
\newblock \emph{Computational Statistics \& Data Analysis}, 55\penalty0
  (12):\penalty0 3232 -- 3243, 2011.
\newblock ISSN 0167-9473.
\newblock \doi{10.1016/j.csda.2011.06.006}.
\newblock URL
  \url{http://www.sciencedirect.com/science/article/pii/S0167947311002076}.

\bibitem[Drechsler \& Shlomo(2018)Drechsler and Shlomo]{Drechsler2018b}
Drechsler, J. and Shlomo, N.
\newblock {Preface to the papers on 'Data confidentiality and statistical
  disclosure control'}.
\newblock \emph{Journal of the Royal Statistical Society: Series A (Statistics
  in Society)}, 181\penalty0 (3):\penalty0 607--608, 2018.
\newblock \doi{10.1111/rssa.12383}.

\bibitem[Duncan \& Stokes(2004)Duncan and Stokes]{Duncan01}
Duncan, G.~T. and Stokes, S.~L.
\newblock {Disclosure Risk vs. Data Utility: The R-U Confidentiality Map as
  Applied to Topcoding}.
\newblock \emph{CHANCE}, 17\penalty0 (3):\penalty0 16--20, 2004.
\newblock \doi{10.1080/09332480.2004.10554908}.

\bibitem[Dwork(2006)]{differential-privacy}
Dwork, C.
\newblock {Differential Privacy}.
\newblock In \emph{33rd International Colloquium on Automata, Languages and
  Programming, part II (ICALP 2006)}, volume 4052, pp.\  1--12, Venice, Italy,
  2006. Springer Verlag.
\newblock ISBN 3-540-35907-9.

\bibitem[Dwork \& Naor(2010)Dwork and Naor]{Dwork2010}
Dwork, C. and Naor, M.
\newblock {On the Difficulties of Disclosure Prevention in Statistical
  Databases or the Case for Differential Privacy}.
\newblock \emph{Journal of Privacy and Confidentiality}, 2\penalty0
  (1):\penalty0 93--107, 2010.
\newblock \doi{10.29012/jpc.v2i1.585}.

\bibitem[Dwork \& Roth(2014)Dwork and Roth]{Dwork2013}
Dwork, C. and Roth, A.
\newblock {The Algorithmic Foundations of Differential Privacy}.
\newblock \emph{Foundations and Trends{\textregistered} in Theoretical Computer
  Science}, 9\penalty0 (3-4):\penalty0 211--407, 2014.
\newblock \doi{10.1561/0400000042}.

\bibitem[Dwork et~al.(2006)Dwork, McSherry, Nissim, and Smith]{Dwork2006}
Dwork, C., McSherry, F., Nissim, K., and Smith, A.
\newblock {Calibrating Noise to Sensitivity in Private Data Analysis}.
\newblock In Halevi, S. and Rabin, T. (eds.), \emph{{Theory of Cryptography}},
  pp.\  265--284, Berlin, Heidelberg, 2006. Springer Berlin Heidelberg.
\newblock ISBN 978-3-540-32732-5.
\newblock \doi{10.1007/11681878_14}.

\bibitem[Dwork et~al.(2017)Dwork, Smith, Steinke, and Ullman]{Dwork2017}
Dwork, C., Smith, A., Steinke, T., and Ullman, J.
\newblock {Exposed! A Survey of Attacks on Private Data}.
\newblock \emph{Annual Review of Statistics and Its Application}, 4\penalty0
  (1):\penalty0 61--84, 2017.
\newblock \doi{10.1146/annurev-statistics-060116-054123}.

\bibitem[Elliot \& Domingo{-}Ferrer(2018)Elliot and
  Domingo{-}Ferrer]{elliot2018}
Elliot, M. and Domingo{-}Ferrer, J.
\newblock The future of statistical disclosure control.
\newblock \emph{Paper published as part of The National Statistician's Quality
  Review}, 2018.

\bibitem[Evans et~al.(Working Paper)Evans, King, Schwenzfeier, and
  Thakurta]{Evans2019}
Evans, G., King, G., Schwenzfeier, M., and Thakurta, A.
\newblock Statistically valid inferences from privacy protected data, Working
  Paper.
\newblock URL \url{https://gking.harvard.edu/dp}.

\bibitem[Fisher(1953)]{Fisher1935}
Fisher, R.~A.
\newblock \emph{The design of experiments}.
\newblock Edinburgh, 6. ed., repr. edition, 1953.

\bibitem[Frigerio et~al.(2019)Frigerio, de~Oliveira, Gomez, and
  Duverger]{FrigerioOGD19}
Frigerio, L., de~Oliveira, A.~S., Gomez, L., and Duverger, P.
\newblock {Differentially Private Generative Adversarial Networks for Time
  Series, Continuous, and Discrete Open Data}.
\newblock In Dhillon, G., Karlsson, F., Hedstr{\"o}m, K., and Z{\'u}quete, A.
  (eds.), \emph{{ICT Systems Security and Privacy Protection}}, pp.\  151--164,
  Cham, 2019. Springer International Publishing.
\newblock ISBN 978-3-030-22312-0.
\newblock \doi{10.1007/978-3-030-22312-0_11}.

\bibitem[Ghosh \& Roth(2015)Ghosh and Roth]{ghosh2015}
Ghosh, A. and Roth, A.
\newblock Selling privacy at auction.
\newblock \emph{Games and Economic Behavior}, 91:\penalty0 334 -- 346, 2015.
\newblock ISSN 0899-8256.
\newblock \doi{10.1016/j.geb.2013.06.013}.
\newblock URL
  \url{http://www.sciencedirect.com/science/article/pii/S0899825613000961}.

\bibitem[Goodfellow et~al.(2014)Goodfellow, Pouget-Abadie, Mirza, Xu,
  Warde-Farley, Ozair, Courville, and Bengio]{Goodfellow2014}
Goodfellow, I., Pouget-Abadie, J., Mirza, M., Xu, B., Warde-Farley, D., Ozair,
  S., Courville, A., and Bengio, Y.
\newblock Generative adversarial nets.
\newblock In Ghahramani, Z., Welling, M., Cortes, C., Lawrence, N., and
  Weinberger, K.~Q. (eds.), \emph{Advances in Neural Information Processing
  Systems}, volume~27, pp.\  2672--2680. Curran Associates, Inc., 2014.
\newblock URL
  \url{https://proceedings.neurips.cc/paper/2014/file/5ca3e9b122f61f8f06494c97b1afccf3-Paper.pdf}.

\bibitem[Hitaj et~al.(2017)Hitaj, Ateniese, and Perez-Cruz]{Hitaj2017}
Hitaj, B., Ateniese, G., and Perez-Cruz, F.
\newblock {Deep Models Under the GAN: Information Leakage from Collaborative
  Deep Learning}.
\newblock In \emph{{Proceedings of the 2017 ACM SIGSAC Conference on Computer
  and Communications Security}}, CCS '17, pp.\  603--618, New York, NY, USA,
  2017. Association for Computing Machinery.
\newblock ISBN 9781450349468.
\newblock \doi{10.1145/3133956.3134012}.
\newblock URL \url{https://doi.org/10.1145/3133956.3134012}.

\bibitem[Huang et~al.(2019)Huang, McKenna, Bissias, Miklau, Hay, and
  Machanavajjhala]{huang2019}
Huang, Z., McKenna, R., Bissias, G., Miklau, G., Hay, M., and Machanavajjhala,
  A.
\newblock {PSynDB: Accurate and Accessible Private Data Generation}.
\newblock \emph{Proc. VLDB Endow.}, 12\penalty0 (12):\penalty0 1918--1921,
  August 2019.
\newblock ISSN 2150-8097.
\newblock \doi{10.14778/3352063.3352099}.
\newblock URL \url{https://doi.org/10.14778/3352063.3352099}.

\bibitem[Hundepool et~al.(2012)Hundepool, Domingo-Ferrer, Franconi, Giessing,
  Nordholt, Spicer, and De~Wolf]{hundepool2012}
Hundepool, A., Domingo-Ferrer, J., Franconi, L., Giessing, S., Nordholt, E.~S.,
  Spicer, K., and De~Wolf, P.-P.
\newblock \emph{{Statistical Disclosure Control}}.
\newblock John Wiley \& Sons, 2012.
\newblock \doi{10.1002/9781118348239}.

\bibitem[Jang et~al.(2017)Jang, Gu, and Poole]{jang2016categorical}
Jang, E., Gu, S., and Poole, B.
\newblock Categorical reparameterization with gumbel-softmax.
\newblock In \emph{{5th International Conference on Learning Representations,
  {ICLR} 2017, Toulon, France, April 24-26, 2017, Conference Track
  Proceedings}}. OpenReview.net, 2017.
\newblock URL \url{https://openreview.net/forum?id=rkE3y85ee}.

\bibitem[Karwa et~al.(2017)Karwa, Krivitsky, and Slavkovi{\'c}]{Karwa2017}
Karwa, V., Krivitsky, P.~N., and Slavkovi{\'c}, A.~B.
\newblock {Sharing Social Network Data: Differentially Private Estimation of
  Exponential Family Random-Graph Models}.
\newblock \emph{Journal of the Royal Statistical Society: Series C (Applied
  Statistics)}, 66\penalty0 (3):\penalty0 481--500, 2017.
\newblock \doi{10.1111/rssc.12185}.

\bibitem[Kinney et~al.(2010)Kinney, Reiter, and Berger]{Kinney2010}
Kinney, S., Reiter, J., and Berger, J.~O.
\newblock {Model Selection When Multiple Imputation is Used to Protect
  Confidentiality in Public Use Data}.
\newblock \emph{Journal of Privacy and Confidentiality}, 2\penalty0
  (2):\penalty0 3--19, 2010.
\newblock \doi{10.29012/jpc.v2i2.588}.

\bibitem[Kinney et~al.(2011)Kinney, Reiter, Reznek, Miranda, Jarmin, and
  Abowd]{Kinney2011}
Kinney, S.~K., Reiter, J.~P., Reznek, A.~P., Miranda, J., Jarmin, R.~S., and
  Abowd, J.~M.
\newblock {Towards Unrestricted Public Use Business Microdata: The Synthetic
  Longitudinal Business Database}.
\newblock \emph{International Statistical Review}, 79\penalty0 (3):\penalty0
  362--384, nov 2011.
\newblock \doi{10.1111/j.1751-5823.2011.00153.x}.

\bibitem[Little(1993)]{Little1993}
Little, R.~J.
\newblock {Statistical Analysis of Masked Data}.
\newblock \emph{Journal of Official Statistics}, 9\penalty0 (2):\penalty0
  407--426, 1993.
\newblock URL
  \url{https://www.scb.se/contentassets/ca21efb41fee47d293bbee5bf7be7fb3/statistical-analysis-of-masked-data.pdf}.

\bibitem[Liu(2017)]{Liu2016}
Liu, F.
\newblock Model-based differentially private data synthesis, 2017.

\bibitem[Loukides et~al.(2012)Loukides, Gkoulalas-Divanis, and
  Shao]{loukides2012}
Loukides, G., Gkoulalas-Divanis, A., and Shao, J.
\newblock Assessing disclosure risk and data utility trade-off in transaction
  data anonymization.
\newblock \emph{International Journal of Software and Informatics}, 6\penalty0
  (3):\penalty0 399--417, 2012.

\bibitem[Maddison et~al.(2017)Maddison, Mnih, and Teh]{maddison2016concrete}
Maddison, C.~J., Mnih, A., and Teh, Y.~W.
\newblock The concrete distribution: {A} continuous relaxation of discrete
  random variables.
\newblock In \emph{{5th International Conference on Learning Representations,
  {ICLR} 2017, Toulon, France, April 24-26, 2017, Conference Track
  Proceedings}}. OpenReview.net, 2017.
\newblock URL \url{https://openreview.net/forum?id=S1jE5L5gl}.

\bibitem[Manrique-Vallier \& Hu(2018)Manrique-Vallier and
  Hu]{Manrique-Vallier2018}
Manrique-Vallier, D. and Hu, J.
\newblock {Bayesian non-parametric generation of fully synthetic multivariate
  categorical data in the presence of structural zeros}.
\newblock \emph{Journal of the Royal Statistical Society: Series A (Statistics
  in Society)}, 181\penalty0 (3):\penalty0 635--647, 2018.
\newblock \doi{10.1111/rssa.12352}.
\newblock URL
  \url{https://rss.onlinelibrary.wiley.com/doi/abs/10.1111/rssa.12352}.

\bibitem[McClure \& Reiter(2012)McClure and Reiter]{McClure2012}
McClure, D. and Reiter, J.~P.
\newblock {Differential Privacy and Statistical Disclosure Risk Measures: An
  Investigation with Binary Synthetic Data}.
\newblock \emph{Trans. Data Privacy}, 5\penalty0 (3):\penalty0 535--552,
  December 2012.
\newblock ISSN 1888-5063.
\newblock \doi{10.5555/2423656.2423658}.

\bibitem[Meng(1994)]{Meng1994}
Meng, X.-L.
\newblock {Multiple-Imputation Inferences with Uncongenial Sources of Input}.
\newblock \emph{Statist. Sci.}, 9\penalty0 (4):\penalty0 538--558, 11 1994.
\newblock \doi{10.1214/ss/1177010269}.
\newblock URL \url{https://doi.org/10.1214/ss/1177010269}.

\bibitem[Mironov(2017)]{Mironov17}
Mironov, I.
\newblock {{R{\'{e}}nyi Differential Privacy}}.
\newblock In \emph{{2017 IEEE 30th Computer Security Foundations Symposium
  (CSF)}}, pp.\  263--275, 2017.
\newblock \doi{10.1109/CSF.2017.11}.

\bibitem[Purdam \& Elliot(2007)Purdam and Elliot]{purdam2007}
Purdam, K. and Elliot, M.
\newblock {A Case Study of the Impact of Statistical Disclosure Control on Data
  Quality in the Individual UK Samples of Anonymised Records}.
\newblock \emph{Environment and Planning A: Economy and Space}, 39\penalty0
  (5):\penalty0 1101--1118, 2007.
\newblock \doi{10.1068/a38335}.

\bibitem[Raghunathan et~al.(2003)Raghunathan, Reiter, and
  Rubin]{Raghunathan2003}
Raghunathan, T.~E., Reiter, J., and Rubin, D.
\newblock {Multiple Imputation for Statistical Disclosure Limitation}.
\newblock \emph{Journal of Official Statistics}, 19\penalty0 (1):\penalty0 1,
  2003.
\newblock URL
  \url{https://www.scb.se/contentassets/ca21efb41fee47d293bbee5bf7be7fb3/multiple-imputation-for-statistical-disclosure-limitation.pdf}.

\bibitem[Reiter(2005{\natexlab{a}})]{Reiter2005}
Reiter, J.
\newblock {Using CART to Generate Partially Synthetic, Public Use Microdata}.
\newblock \emph{Journal of Official Statistics}, 21\penalty0 (3):\penalty0
  441--462, 2005{\natexlab{a}}.
\newblock URL
  \url{https://www.scb.se/contentassets/ca21efb41fee47d293bbee5bf7be7fb3/using-cart-to-generate-partially-synthetic-public-use-microdata.pdf}.

\bibitem[Reiter(2005{\natexlab{b}})]{reiter2005b}
Reiter, J.
\newblock {Estimating Risks of Identification Disclosure in Microdata}.
\newblock \emph{Journal of the American Statistical Association}, 100\penalty0
  (472):\penalty0 1103--1112, 2005{\natexlab{b}}.
\newblock \doi{10.1198/016214505000000619}.

\bibitem[Reiter(2005{\natexlab{c}})]{reiter2005c}
Reiter, J.
\newblock Releasing multiply imputed, synthetic public use microdata: an
  illustration and empirical study.
\newblock \emph{Journal of the Royal Statistical Society: Series A (Statistics
  in Society)}, 168\penalty0 (1):\penalty0 185--205, 2005{\natexlab{c}}.
\newblock \doi{10.1111/j.1467-985X.2004.00343.x}.

\bibitem[Reiter \& Raghunathan(2007)Reiter and Raghunathan]{Reiter2007}
Reiter, J. and Raghunathan, T.~E.
\newblock {The multiple adaptations of multiple imputation}.
\newblock \emph{Journal of the American Statistical Association}, 102\penalty0
  (480):\penalty0 1462--1471, 2007.
\newblock \doi{10.1198/016214507000000932}.

\bibitem[Robinson \& Turner(2017)Robinson and Turner]{Robinson2017}
Robinson, A. and Turner, K.
\newblock {Hypothesis Testing for Topological Data Analysis}.
\newblock \emph{Journal of Applied and Computational Topology}, 1\penalty0
  (2):\penalty0 241--261, 2017.
\newblock \doi{10.1007/s41468-017-0008-7}.

\bibitem[Rubin(1993)]{Rubin1993}
Rubin, D.
\newblock {Discussion: Statistical Disclosure Limitation}.
\newblock \emph{Journal of Official Statistics}, 9\penalty0 (2):\penalty0
  461--468, 1993.
\newblock URL
  \url{https://www.scb.se/contentassets/ca21efb41fee47d293bbee5bf7be7fb3/discussion-statistical-disclosure-limitation2.pdf}.

\bibitem[Rubin(1996)]{Rubin1996}
Rubin, D.
\newblock {Multiple Imputation after 18+ Years}.
\newblock \emph{Journal of the American Statistical Association}, 91\penalty0
  (434):\penalty0 473--489, 1996.
\newblock \doi{10.1080/01621459.1996.10476908}.

\bibitem[Shlomo(2018)]{Shlomo2018}
Shlomo, N.
\newblock {Statistical Disclosure Limitation: New Directions and Challenges}.
\newblock \emph{Journal of Privacy and Confidentiality}, 8\penalty0 (1), 2018.
\newblock \doi{10.29012/jpc.684}.

\bibitem[Shokri et~al.(2016)Shokri, Theodorakopoulos, and Troncoso]{shokri2016}
Shokri, R., Theodorakopoulos, G., and Troncoso, C.
\newblock {Privacy Games Along Location Traces: A Game-Theoretic Framework for
  Optimizing Location Privacy}.
\newblock \emph{ACM Trans. Priv. Secur.}, 19\penalty0 (4), December 2016.
\newblock ISSN 2471-2566.
\newblock \doi{10.1145/3009908}.
\newblock URL \url{https://doi.org/10.1145/3009908}.

\bibitem[Snoke et~al.(2018)Snoke, Raab, Nowok, Dibben, and
  Slavkovic]{Snoke2018}
Snoke, J., Raab, G.~M., Nowok, B., Dibben, C., and Slavkovic, A.
\newblock General and specific utility measures for synthetic data.
\newblock \emph{Journal of the Royal Statistical Society: Series A (Statistics
  in Society)}, 181\penalty0 (3):\penalty0 663--688, 2018.
\newblock \doi{https://doi.org/10.1111/rssa.12358}.

\bibitem[Taylor et~al.(2018)Taylor, Zhou, and Rise]{taylor2018}
Taylor, L., Zhou, X.-H., and Rise, P.
\newblock A tutorial in assessing disclosure risk in microdata.
\newblock \emph{Statistics in Medicine}, 37\penalty0 (25):\penalty0 3693--3706,
  2018.
\newblock \doi{10.1002/sim.7667}.

\bibitem[Torkzadehmahani et~al.(2019)Torkzadehmahani, Kairouz, and
  Paten]{TKP20}
Torkzadehmahani, R., Kairouz, P., and Paten, B.
\newblock {DP-CGAN: Differentially Private Synthetic Data and Label
  Generation}.
\newblock In \emph{{2019 IEEE/CVF Conference on Computer Vision and Pattern
  Recognition Workshops (CVPRW)}}, pp.\  98--104, 2019.
\newblock \doi{10.1109/CVPRW.2019.00018}.

\bibitem[Tripathy et~al.(2019)Tripathy, Wang, and Ishwar]{Tripathy2017}
Tripathy, A., Wang, Y., and Ishwar, P.
\newblock {Privacy-Preserving Adversarial Networks}.
\newblock In \emph{{2019 57th Annual Allerton Conference on Communication,
  Control, and Computing (Allerton)}}, pp.\  495--505, 2019.
\newblock \doi{10.1109/ALLERTON.2019.8919758}.

\bibitem[Trottini \& Fienberg(2002)Trottini and Fienberg]{trottini2002}
Trottini, M. and Fienberg, S.~E.
\newblock Modelling user uncertainty for disclosure risk and data utility.
\newblock \emph{Int. J. Uncertain. Fuzziness Knowl.-Based Syst.}, 10\penalty0
  (5):\penalty0 511--527, October 2002.
\newblock ISSN 0218-4885.
\newblock \doi{10.1142/S0218488502001612}.
\newblock URL \url{https://doi.org/10.1142/S0218488502001612}.

\bibitem[Van~Buuren(2018)]{vanBuuren2018}
Van~Buuren, S.
\newblock \emph{Flexible Imputation of Missing Data}.
\newblock Chapman and Hall/CRC, 2018.
\newblock \doi{10.1201/9780429492259}.

\bibitem[Woo et~al.(2009)Woo, Reiter, Oganian, and Karr]{Woo2009}
Woo, M.-J., Reiter, J., Oganian, A., and Karr, A.~F.
\newblock {Global Measures of Data Utility for Microdata Masked for Disclosure
  Limitation}.
\newblock \emph{Journal of Privacy and Confidentiality}, 1\penalty0
  (1):\penalty0 111--124, 2009.
\newblock \doi{10.29012/jpc.v1i1.568}.

\bibitem[Wood et~al.(2018)Wood, Altman, Bembenek, Bun, Gaboardi, Honaker,
  Nissim, OBrien, Steinke, and Vadhan]{Nissim2017}
Wood, A., Altman, M., Bembenek, A., Bun, M., Gaboardi, M., Honaker, J., Nissim,
  K., OBrien, D.~R., Steinke, T., and Vadhan, S.
\newblock {Differential privacy: A primer for a non-technical audience}.
\newblock \emph{Vanderbilt Journal of Entertainment \& Technology Law},
  21\penalty0 (1):\penalty0 209--275, 2018.
\newblock URL
  \url{https://salil.seas.harvard.edu/files/salil/files/differential_privacy_primer_nontechnical_audience.pdf}.

\bibitem[Wu et~al.(2019)Wu, Yang, Xu, and Ling]{Wu2018}
Wu, Y., Yang, F., Xu, Y., and Ling, H.
\newblock {Privacy-Protective-GAN for Privacy Preserving Face
  De-Identification}.
\newblock \emph{Journal of Computer Science and Technology}, 34\penalty0
  (1):\penalty0 47--60, Jan 2019.
\newblock ISSN 1860-4749.
\newblock \doi{10.1007/s11390-019-1898-8}.
\newblock URL \url{https://doi.org/10.1007/s11390-019-1898-8}.

\bibitem[Xie et~al.(2018)Xie, Lin, Wang, Wang, and Zhou]{Xie2018}
Xie, L., Lin, K., Wang, S., Wang, F., and Zhou, J.
\newblock {Differentially Private Generative Adversarial Network}, 2018.

\bibitem[Xie \& Meng(2017)Xie and Meng]{Xie2017}
Xie, X. and Meng, X.~L.
\newblock {Dissecting Multiple Imputation From a Multi-Phase Inference
  Perspective: What Happens When God's, Imputer's and Analyst's Models Are
  Uncongenial?}
\newblock \emph{Statistica Sinica}, 27\penalty0 (4):\penalty0 1485--1545, 2017.
\newblock \doi{10.5705/ss.2014.067}.

\bibitem[Xu et~al.(2019)Xu, Ren, Zhang, Zhang, Qin, and Ren]{Xu2019}
Xu, C., Ren, J., Zhang, D., Zhang, Y., Qin, Z., and Ren, K.
\newblock {GANobfuscator: Mitigating Information Leakage Under GAN via
  Differential Privacy}.
\newblock \emph{IEEE Transactions on Information Forensics and Security},
  14\penalty0 (9):\penalty0 2358--2371, 2019.
\newblock \doi{10.1109/TIFS.2019.2897874}.

\bibitem[Zhang et~al.(2017)Zhang, Cormode, Procopiuc, Srivastava, and
  Xiao]{Zhang2017}
Zhang, J., Cormode, G., Procopiuc, C.~M., Srivastava, D., and Xiao, X.
\newblock {PrivBayes: Private Data Release via Bayesian Networks}.
\newblock \emph{ACM Trans. Database Syst.}, 42\penalty0 (4), October 2017.
\newblock ISSN 0362-5915.
\newblock \doi{10.1145/3134428}.
\newblock URL \url{https://doi.org/10.1145/3134428}.

\end{thebibliography}
\bibliographystyle{icml2019}
\newpage
\appendix
\section{Appendix}

\subsection{A Brief Introduction to GANs.}

The basic idea of a GAN is surprisingly intuitive. At its core, a GAN is a minimax game with two competing actors. A discriminator (D) tries to tell real from synthetic samples and a generator (G) intends to produce realistic synthetic samples from random noise. We use the same illustrative example as \citet{Goodfellow2014} to make GANs (and the adjustments later on) more accessible: ``The generative model can be thought of as analogous to a team of counterfeiters, trying to produce fake currency and use it without detection, while the discriminative model is analogous to the police, trying to detect the counterfeit currency. Competition in this game drives both teams to improve their methods until the counterfeits are indistinguishable from the genuine articles.''

In a GAN, the team of counterfeiters (Generator G) is a neural network which is trained to produce realistic synthetic data examples from random noise. The police (discriminator D) is a neural network with the goal to distinguish fake data from real data. The generator network trains to fool the discriminator network. It uses the feedback of the discriminator to generate increasingly realistic ``fake'' data that should eventually be indistinguishable from the original ones. At the same time, the discriminator is constantly adapting to the more and more improving generating abilities of the generator. Thus, the ``threshold'' where the discriminator is fooled increases along with the generator's capability to convincingly generate data similar to the original data. This goes on until equilibrium is reached\footnote{Interestingly, a GAN is therefore a dynamic system where the optimisation process is not seeking a minimum, but an equilibrium instead. This is in stark contrast to standard deep learning systems, where the entire loss landscape is static.}.

Formally, this two-player minimax game can be written as:
\begin{align*}
\min_{G} \max_{D} V(D,G) = &\mathbb{E}_{x \sim p_{\rm data}(x)}\Big[ \log D(x)\Big] + \\ & \mathbb{E}_{z \sim p_{\rm z}(z)}\Big[\log (1-D(G(z)))\Big]
\end{align*}

where $p_{data}(x)$ is the distribution of the real data, $X$ is a sample from $p_{data}(x)$.
The generator network $G(z)$ takes as input $z$ from $p(z)$, where $z$ is a random sample from a probability distribution $p(z)$\footnote{Usually GANs are set up to either sample from uniform or Gaussian distributions.}. Passing the noise $z$ through $G$ then generates a sample of synthetic data feeding the discriminator $D(x)$. The discriminator receives as input a set of labeled data, either real $(x)$ from $p_{data}(x)$ or generated $(G(z))$, and is trained to distinguish between real data and synthetic data\footnote{This is a standard binary classification problem, and thus the standard binary cross-entropy loss with a sigmoid function at the end can be used.}. $D$ is trained to maximize the probability of assigning the correct label to training examples and samples from $G(z)$. $G$ is trained to minimize $log(1 - D(G(z)))$. Ultimately, the goal of the discriminator is to maximize function $V$, whereas the goal of the generator is to minimize it.

The equilibrium point for the GANs is that the $G$ should model the real data and $D$ should output the probability of 0.5 as the generated data is same as the real data -- that is, it is not sure if the new data coming from the generator is real or fake with equal probability.\footnote{Note the connection to the measure of general utility presented by \citet{Snoke2018}. The explicit goal of a GAN is to maximize general utility, and therefore a natural way to generate fully synthetic data.}

\subsection{A Data Generating Process for Scenario 1.}

\begin{align}
    x_1 &\sim \mathcal{N}(5, 2) \label{eq:x1}\\
    x_2 &\sim \mathcal{N}(-3, 1) \label{eq:x1b}\\
    x_3 &\sim \mathcal{BN}(1, 0.7) \label{eq:x2}\\
    x_4 &\sim \mathcal{NB}(0.8, 30) \label{eq:x3}\\
    x_5 &\sim \mathcal{MN}(1, 0.2, 0.3, 0.5) \label{eq:x4}\\
    x_6 &\sim \mathcal{N}(x_3, 50) \label{eq:x5}\\
    w_1 &\sim \mathcal{BN}(1, 0.5) \label{eq:w1}\\
    w_2 &\sim \mathcal{BN}(1, 0.3 w_1) \label{eq:w2}\\
    y &\sim \mathcal{N}(x_1 + x_2 + x_3 + x_4 + x_1 * x_4, 20) \label{eq:epsilon}\\
\end{align}

\end{document}